\documentclass[10pt,twocolumn,letterpaper]{article}

\usepackage[pagenumbers]{utils/cvpr} 

\usepackage{xcolor}         
\definecolor{myBlue}{HTML}{DEECF4}

\usepackage{multirow}
\usepackage{tabularx, booktabs, multirow, makecell, xcolor, colortbl}
\usepackage{amssymb,pifont}
\definecolor{headergray}{RGB}{248,248,248} 
\usepackage{tabularx, booktabs, makecell, array, xcolor, colortbl}
\newcolumntype{R}{>{\raggedleft\arraybackslash}X}  
\definecolor{grpgray}{gray}{0.96}                  

\DeclareMathOperator*{\argmin}{arg\,min}

\definecolor{duetD}{HTML}{5B4B8A}  
\definecolor{duetU}{HTML}{7B6BA8}  
\definecolor{duetE}{HTML}{9B8BC6}  
\definecolor{duetT}{HTML}{B8AAD9}  
\definecolor{duetG}{HTML}{E8956B}  
\definecolor{duetE2}{HTML}{EA7A52} 
\definecolor{duetN}{HTML}{E65C3B}  

\newcommand{\method}{RDA\xspace}

\definecolor{mygray}{gray}{0.5}

\definecolor{myblack}{RGB}{40,40,40}

\definecolor{mygreen}{RGB}{0,110,150}

\newcommand{\dualcolor}[2]{%
  \textcolor{mygray}{#1}/\textcolor{duetD}{{#2}}%
}

\usepackage{tcolorbox}
\tcbuselibrary{listings,breakable}

\definecolor{citecolor}{RGB}{0, 90, 160}
\newtcolorbox{prompt}[2][]{
  colback=citecolor!6!white,        
  title={#2}, #1}

\definecolor{cvprblue}{rgb}{0.21,0.49,0.74}
\usepackage[pagebackref,breaklinks,colorlinks,allcolors=cvprblue]{hyperref}

\def\paperID{3014} 
\def\confName{CVPR}
\def\confYear{2026}

\title{Residual Decoder Adapter:\\ ID-Preserving Tokenizer Adaption for Autoregressive Text Rendering}

\author{Dongxing Mao$^{1\star}$, Alex Jinpeng Wang$^{1\star}$, Jiahao Tang$^{1}$, Kevin Qinghong Lin$^{2}$, \\
Linjie Li$^{3}$, Zhengyuan Yang$^{3}$, Lijuan Wang$^{3}$, Min Li$^{1}$, Jingru Tan$^{1\dagger}$\\
	\\
	$^{1}$Central South University\quad$^{2}$University of Oxford\quad
	$^{3}$Microsoft Research
}

\begin{document}
\maketitle

\let\thefootnote\relax\footnotetext{$^\star$ Equal contribution.} 
\footnotetext[2]{$\dagger$ Corresponding author }
\begin{abstract}
Visual Autoregressive (AR) models generate images by predicting discrete tokens that are decoded by a visual tokenizer.
Despite demonstrating strong overall image generation ability, they still underperform on text rendering with blur strokes and disrupt letter shapes. 
In this work, we trace this limitation to the visual tokenizer, which struggles to reconstruct fine-grained detail.
Improving the tokenizer is straightforward but expensive, as it necessitates retraining both the tokenizer and the AR model.
Can we improve text rendering performance of AR models without retraining the existing tokenizer and AR model? 
To achieve this, we propose the Residual Decoder Adapter(\method) that upgrades an existing tokenizer post-hoc without changing its token space. 
Specifically, it refines the decoder output of the visual tokenizer by introducing two novel components: 
(i) a paired codebook that shares the token distribution with the original one; 
(ii) a parallel branch to learn the tiny differences (residual) between the reconstructed image and the ground-truth images in the pixel space. This residual design allows us to enhance the tokenizer non-invasively while preserving compatibility with prior AR models. \method substantially improves text rendering significantly by a large margin. 
For instance, we boost finetuned Janus-Pro OCR accuracy rises from 24.52\% to 58.26\% (TextVisionBlend), from 12.75\% to 36.81\% (StyledTextSynth) on competitive TextAtlas benchmark. The code is available at \href{https://github.com/CSU-JPG/RDA}{github.com/CSU-JPG/RDA}.

\end{abstract}    
\section{Introduction}
\label{sec:intro}

\begin{figure}[t]
  \centering
  \includegraphics[width=0.48\textwidth]{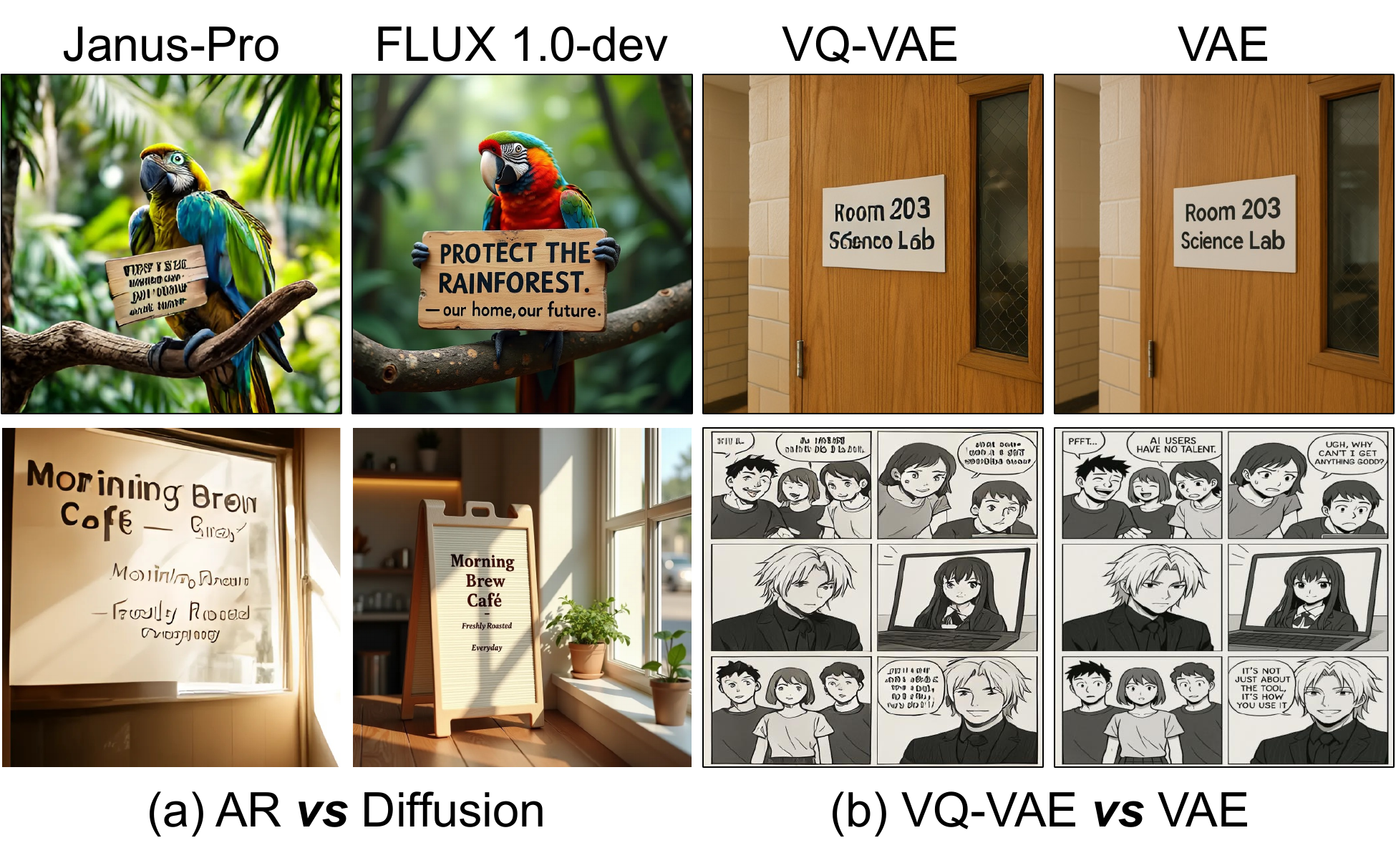} 
  \caption{(a) \textbf{Comparison between AR model (\textit{Janus-Pro}) and Diffusion model (\textit{FLUX 1.0-dev})}. They exhibit similar performance on the general text-to-image benchmark GenEval, with accuracy of 0.80 and 0.82, respectively. (b) Comparison between VQ-VAE (from Janus-Pro) and VAE (from FLUX) at 512 resolution. Their reconstruction metrics on ImageNet with rFID scores of 9.63 and 7.92, respectively. }
  \label{fig:problem_illustration}
\end{figure}

Visual Autoregressive (AR) models~\cite{chameleon,januspro,VAR} have emerged as a powerful paradigm for image generation, predicting discrete \textit{visual tokens} that are decoded by a visual tokenizer~\cite{vqvae,vqgan}. 
Despite competitive performance on standard benchmarks, AR models face a significant challenge that has recently drawn considerable attention in the community: text rendering. This task has become widely recognized as one of the most demanding tests for fine-grained generation capability~\cite{emu3, bagel,gpt4o,wu2025qwenimagetechnicalreport}, yet AR models substantially lag behind diffusion models~\cite{sd,flux}, producing blurred strokes and distorted shapes (~\cref{fig:problem_illustration}(a)).

The weak text rendering of AR models largely stems from the visual tokenizer, whose limited reconstruction capability results in the loss of fine-grained textual details (\cref{fig:problem_illustration}(b)), as also observed in recent analyses \cite{vtbench, wang2025wordsadvancinglongtextimage}.

As AR models decode solely through discrete visual tokens, details omitted during tokenization are inherently difficult to recover in subsequent generation.

\begin{figure}[t]
  \centering
  \includegraphics[width=0.48\textwidth]{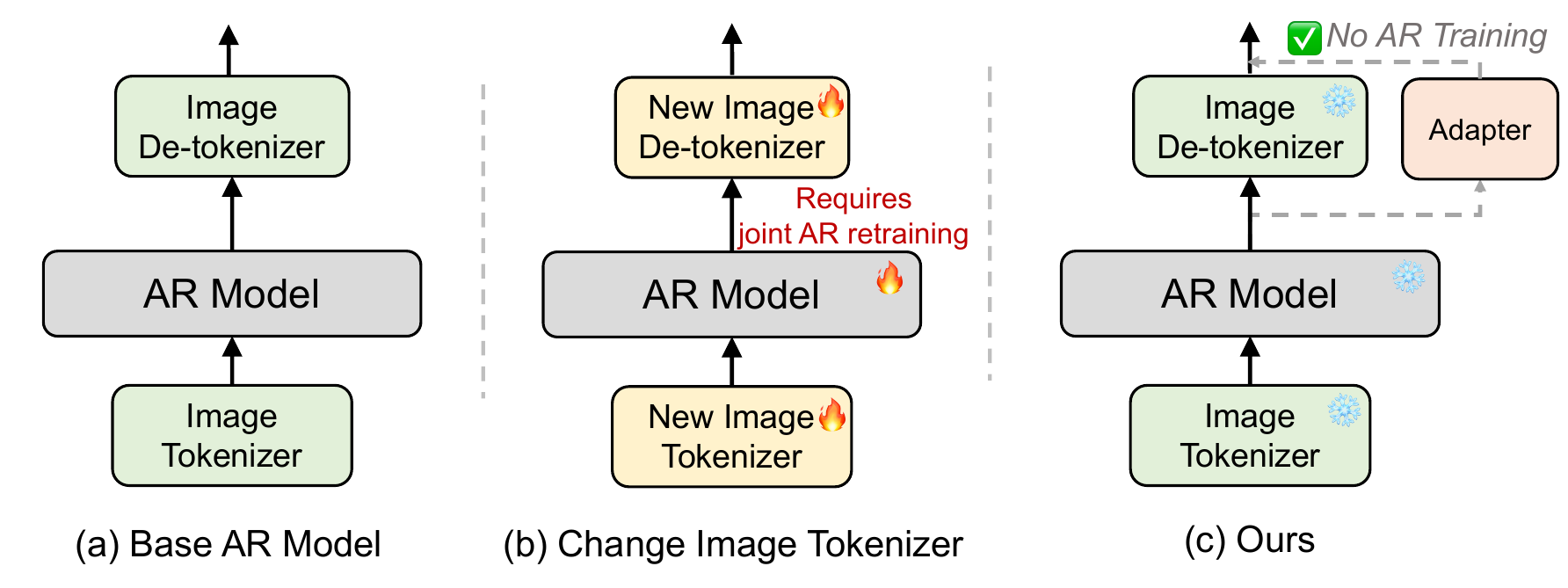} 
  \caption{\textbf{The intuition behind our method. }(a) The modern text-to-image generation ecosystem consists of an AR model and a visual tokenizer. (b) Existing methods develop an improved tokenizer so inevitably re-train AR models. (c) Our approach enhances the tokenizer with an adopter while preserving the compatibility with original AR model(AR-Training Free).}
  \label{fig:intuition_of_our_method}
\end{figure}

Because the tokenizer defines the entire visual language that the AR model can speak, any advance in text rendering must begin with the tokenizer. One strightforward approach is to curate more high-quality data or degsin a sophisticated model to obtain a stronger tokenizer~\cite{qu2025tokenflowunifiedimagetokenizer,xiong2025gigatokscalingvisualtokenizers, unitok}. However, this process incurs significant training costs, since changing token IDs invalidates prior training and requires retraining the entire AR model(\cref{fig:intuition_of_our_method}(b)).

\textit{Can we improve AR models' performance in text rendering without modifying the trained tokenizer and AR model?} Here comes our solution (\cref{fig:intuition_of_our_method}(c)): Instead of change the tokenizer, we attach an plug-in module that refines the outputs of decoder \textbf{without altering its ID space.}

In this work, we propose \textbf{Residual Decoder Adapter (\method)}, a plug-and-play refinement framework that upgrades any pretrained discrete visual tokenizer post-hoc.
\method keeps the entire visual tokenizer frozen and learns two lightweight components:
(1) a learnable Hint Codebook that shares token IDs with the frozen one but provides complementary high-frequency embeddings;
(2) a Residual Decoder that reconstructs fine residual in the pixel space based on the original decoder outputs.
Together, they act as a “visual fidelity booster”, enhancing text clarity while preserving compatibility with downstream AR models.

We conducted extensive experiments. Our method consistently improves both general AR models~(Janus-Pro\cite{januspro}, TAR\cite{tar})~and finetuned text-specific AR models~(Janus-Pro\cite{januspro}, Lumina-mGPT\cite{lumina_mgpt})~on text rendering performance across multiple benchmarks. For instance, applying our method to the  text-specific Janus Pro 1B model increases its text accuracy from 24.52\% to 58.26\% on StyledTextVisionBlend\cite{wang2025textatlas5mlargescaledatasetdense}. Meanwhile, \method enhances text reconstruction capability of the tokenizer while preserving overall image fidelity. On AnyText-Benchmark\cite{tuo2024anytextmultilingualvisualtext}, it improves LlamaGen-VQ\cite{llamagen}  text accuracy from 21.26\% to 36.79\% and reduces LPIPS from 0.1912 to 0.1832. Our design enables scalable visual enhancement without touching the base AR model or retraining the tokenizer.
By reinterpreting the tokenizer not as a static bottleneck but as an extendable interface, \method transforms a long-standing limitation into a pathway for the continual evolution of visual tokenizers. Our contributions are summarized as follows:

\emph{i.} We propose a novel paradigm that bridges the long-standing gap between \textit{discrete VQ tokenizers} and the \textit{continuous image space} in difficult text rendering. 
\method works as an \textit{external tool} that can improve the base AR model without ever \textit{"seeing"} it.
\emph{ii.} Our method extend to mainstream AR models, including Janus-Pro, TAR, and Lumina-mgpt, which employ discrete VQ-based decoders. 
We further show the consistent performance gains over popular benchmarks~\cite{tuo2024anytextmultilingualvisualtext,chen2023textdiffuserdiffusionmodelstext,longtextbench,du2025textcrafteraccuratelyrenderingmultiple,wang2025textatlas5mlargescaledatasetdense}.

\section{Related Work}

\paragraph{Visual Autoregressive Models} 
formulate image synthesis as a next-token prediction task. Pioneering works~\cite{vqvae,vqgan} discretized pixel space into visual tokens, enabling transformers~\cite{transformer} to treat images as sequences of tokens. ~\cite{dalle,cogview} scaled this approach, showing that large-scale AR transformers could generate high-quality, semantically consistent images from text prompts. Recent advancements~\cite{llamagen, dalle, chameleon, janus, januspro, parmar2018image,RAR,ART} have scaled AR models to billion-parameter vision-language models, demonstrating improved compositional reasoning and cross-modal understanding\cite{zhang2023regularized, zheng2022movq}.

Compared to diffusion models~\cite{imagen,sd, ddpm,ddim, xie2024sanaefficienthighresolutionimage,xie2025sana15efficientscaling}, AR models align naturally with large language models~\cite{tong2024metamorphmultimodalunderstandinggeneration}, providing a unified multimodal framework with strong prompt-following and in-context learning~\cite{InstaManip, styleAR}, reasoning and planning abilities~\cite{plangen,januspror1,bagel,textground}, making them ideal for complicate applications like multimodal dialogue~\cite{lumina_mgpt,xin2025luminamgpt20standaloneautoregressive}. However, their visual fidelity remains limited~\cite{lee2022draft,luo2024open}, as performance is bottlenecked by the discrete tokenizer: even with perfect token prediction, reconstruction quality depends on the decoder. This leads to notable degradation in tokenizer-sensitive tasks such as text rendering.
Finally, several recent approaches replace the decoder entirely with a diffusion-based decoder or a DiT head~\cite{d_ar,qwenimage,emu3,bilp3o,dit}, achieving stronger decoding capacity. 
Yet these methods still require joint AR–decoder training and large-scale data, resulting in substantial retraining cost.

\begin{figure*}[tb]
    \centering
    \includegraphics[width=1\linewidth]{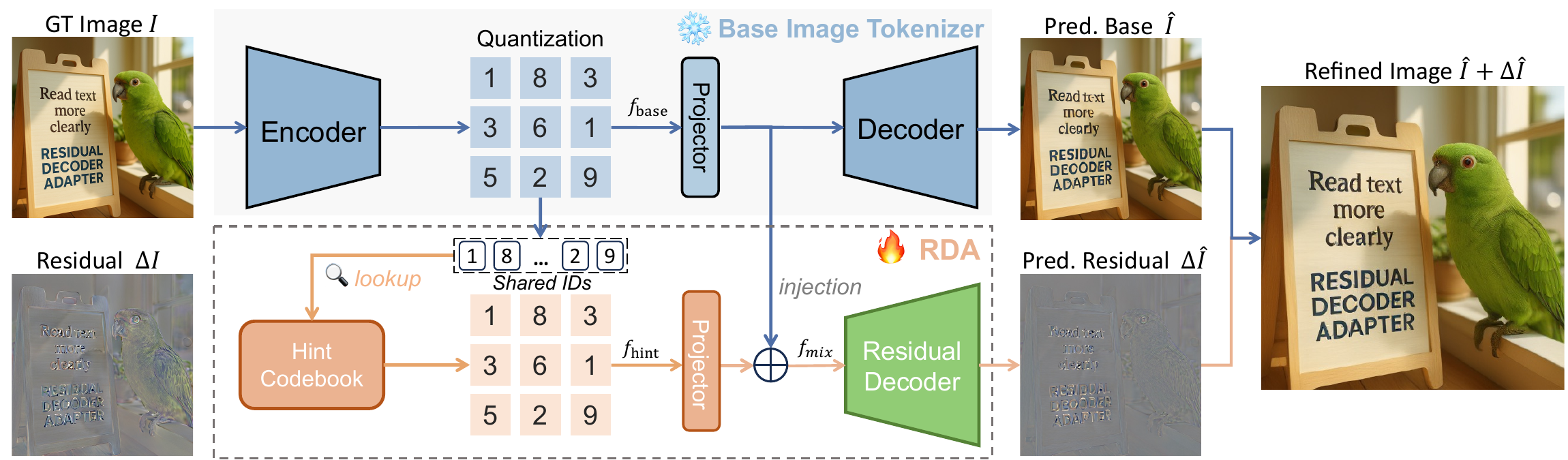}
    \caption{\textbf{Overview of the proposed Residual Decoder Adapter (RDA).}
    RDA enhances an existing image tokenizer without modifying its token space
    or retraining. RDA predict an $\Delta\hat{I}$, which is added into  $\hat{I}$ to obtain the refined output
    $\hat{I} + \Delta\hat{I}$ with more accurate text details.}

    \label{fig:method}
\end{figure*}



%
\paragraph{Visual Tokenizers} plays a key role in autoregressive image generation by quantizing image features into a set of indices~\cite{vqvae,vqgan,dalle}. This process enables Transformers to take images as token sequences, facilitating unified sequence learning across visual and textual modalities. However, quantization inherently introduces reconstruction errors~\cite{vqvae_2}, limiting the visual fidelity of the tokenized representation. This limitation can make the visual tokenizer a significant bottleneck for downstream tasks\cite{chen2025softvq, he2025flowtok,cao2023efficient}.

To enhance reconstruction quality, \cite{vqvae_2} enlarges the codebook to improve representation diversity, \cite{rqvae,maskgit} employs hierarchical quantization to capture multi-scale semantics, \cite{vitvqgan} introduces semantically guided objectives for better perceptual alignment, and \cite{vila-u,unitok} designs a modality-aware tokenizer that unifies visual and textual spaces. \cite{rear} further enhances image tokenizer through token-wise regularization.
However, retraining the tokenizer to modify the codebook alters the distribution of code indices, making it incompatible with the existing trained AR model. 
In contrast, our method decouples reconstruction enhancement from token distribution, allowing direct post-hoc upgrades.

\section{Preliminaries}
\label{sec:preliminaries}

Visual autoregressive generation is divided into two components: (i) A visual tokenizer that deiscretizes continuous images into token sequences; (ii) An autoregressive model that learns the distribution over these tokens.

\noindent\textbf{Visual Tokenizer}. A tokenizer consists of an encoder $\mathcal{E}$, a quantizer $\mathcal{Q}$, and a decoder $\mathcal{D}$, which together transform an image $I$ into a reconstructed version $\hat{{I}}$ via
$\hat{\mathbf{z}}=\mathcal{E}({I}),
\mathbf{z}^q=\mathcal{Q}(\hat{\textbf{z}}),
\hat{{I}}=\mathcal{D}(\mathbf{z}^q)$.

The vector quantization is performed \textit{element-wise} with a codebook $\mathcal{Z} = \{\,\mathbf{z}_1, \mathbf{z}_2, \dots, \mathbf{z}_K\}\;\subset\;\mathbb{R}^{c\times h\times w}$ by looking up the closest entry. Formally:

\begin{equation}
\mathbf{z}_{ij}^q = \mathcal{Q}(\hat{\mathbf{z}}_{ij}), \quad
\mathbf{x}_{ij} = \argmin_{k} \|\hat{\mathbf{z}}_{ij} - \mathbf{z}_k\|.
\end{equation}

Each $\mathbf{x}_{ij}$ serves as the discrete token id.
These ids are then flattened into a sequence  
for training the AR model.

\noindent\textbf{Autoregressive Model.} 
Given a sequence of discrete tokens $\mathbf{x}_{1:N}=\{\mathbf{x}_1,\mathbf{x}_2,\dots,\mathbf{x}_N\}$, 
the AR model $p_\theta$ factorizes the joint distribution as
\begin{equation}
    p_\theta(\mathbf{x}_{1:N}) = \prod_{i=1}^{N} p_\theta(\mathbf{x}_i \mid \mathbf{x}_{1:i-1}),
\end{equation}

and is trained to maximize the log-likelihood $\sum_i\log p_\theta(\mathbf{x}_i|\mathbf{x}_{1:i-1})$. At inference time, visual tokens are generated sequentially and decoded back to an image by decoder $\mathcal{D}$.

\section{Residual Decoder Adapter}

Residual Decoder Adapter (RDA) is plug-and-play framework that improves text rendering quality for visual tokenizers and downstream AR models, while keeping the original tokenizer’s ID space unchanged. The overall architecture is illustrated in \cref{fig:method}. It consists two components: (i) 
a Shared-ID Hint Codebook that provides high-frequency cues, in \cref{sec: hintcodebook}; (ii) a Residual Decoder that refines pixel-level outputs to enhance visual fidelity, in \cref{sec: residual decoder}.

\subsection{Shared-ID Hint Codebook}
\label{sec: hintcodebook}

The quantization in a VQ-VAE maps continuous features to discrete visual codes, which inevitably suppresses high-frequency and local textures, thereby reducing reconstruction fidelity. To compensate for this quantization-induced loss, we introduce a paired \textbf{Hint Codebook} that supplies complementary, detail-oriented features.

Crucially, the Hint Codebook shares the \textbf{same indices} as the original codebook (Shared-ID). In other words, given any token id output by the tokenizer (or from the AR model predictions), we use the \emph{same} id to retrieve features from both codebooks. Specifically, given token id \(i\in\{1,\dots,K\}\) quantized by the tokenizer or predicted by AR models, we retrieve aligned features using the \emph{same} id:

\begin{equation}
\label{label: codebook}
  f_{\text{base}}(i)=z_i, \qquad
  f_{\text{hint}}(i)=z'_i.
\end{equation}

\noindent Here, \(Z\) is \textbf{frozen} codebook \(Z=\{z_k\}_{k=1}^K\)  and \(Z'\) is \textbf{trainable} hint codebook \(Z'=\{z'_k\}_{k=1}^K\). The base features \(f_{\text{base}}\) preserve global structure and semantics; the paired hint features \(f_{\text{hint}}\) provide aligned high-frequency/detail cues.

The Shared-ID design and retrieval rule imply that the two codebooks induce the \emph{same} discrete partition (distribution) over the feature space. Consequently, \(Z'\) acts as an \emph{in-distribution} supplement to \(Z\): for each token id (semantic cluster),\(Z'\)~learns complementary high-frequency details without altering the underlying semantics.

Because original codebook~\(Z\)~is frozen, the mapping from \emph{token id} to the \emph{image distribution} remains unchanged. Therefore, the pretrained AR model, which has learned to predict the original token ID distribution, can directly \emph{plug-and-play} benefit from the improved reconstruction enabled by \(Z'\) without any retraining.

\subsection{Residual Decoder}
\label{sec: residual decoder}



We define the visual detail lost by the tokenizer as the \emph{fine-grained difference} $\Delta I = I - \hat{I}$ between the ground truth image $I$ and the tokenizer-reconstructed image $\hat{I}$.

Given the base features $f_{\text{base}}$ and hint features $f_{\text{hint}}$, our residual decoder $\mathcal{D}_{res}$ predicts $\Delta I$ to recover fine-grained, high-frequency details. 
Overall, the learning consists of two steps: i) Instance-dependent feature injection; ii) Pixel-level residual learning.

\noindent\textbf{Instance-dependent feature injection}. We pass the base and hint features through two projector and fuse them:

\begin{equation}
\label{label: fusion}
  f_{\text{mix}} = p(f_{\text{base}}) + q(f_{\text{hint}})
\end{equation}

Where $p(\cdot)$ is the projector provided by the original tokenizer, $q(\cdot)$ is randomly initialized and trained from scratch. This instance-dependent injection mechanism ensures that the final mixed features carry both the \emph{fine-grained clues} from the hint codebook and the \emph{instance-specific} information from the base branch. This design is critical: removing the instance-dependent injection fails the training of the hint codebook as shown in \cref{tab:feat_add_table}. 

\noindent\textbf{Pixel-level residual learning.} We adopt a parallel, randomly initialized residual decoder \(\mathcal{D}_{\text{res}}\)
predicts the pixel-level residual from the mixed features.

\begin{equation}
    \Delta \hat{I} = \mathcal{D}_{res}(f_{\text{mix}})
\end{equation}

The residual decoder is architecture-agnostic and can be implemented with various neural network designs.
In our implementation, we adopt the VQ-VAE decoder architecture, doubling the channels of the last two convolution layers to better capture high-frequency details.
The final reconstructed image is obtained as:
\begin{equation}
\hat{I}_{\text{final}} =  \hat{I} + \Delta \hat{I}
\end{equation}

\subsection{Training Objective and Optimization}
Although the overall architecture is conceptually simple, taming RDA training is not a trivial task. The optimization difficulties lie in the sparse residual signal, which can be dominated by the background pixels. To ensure successful training of RDA, we 
introduce several loss functions:

\noindent\textbf{Standard losses}. We respectively apply MAE and MSE reconstruction losses to supervise the residual output and the final reconstruction, and denote their combination as $\mathcal{L}_{\text{rec}}$.

Following VQ-VAE~\cite{vqvae}, we also apply perceptual loss on the final reconstructed image by using $\phi(\cdot)$ to extract multi-layer VGG features~\cite{vgg} :
\begin{equation}
    \mathcal{L}^{\text{final}}_{\text{perc}} = \|\phi(I) - \phi(\hat{I}_{\text{final}})\|_2^2
\end{equation}

\noindent\textbf{Residual perceptual loss.}  We also apply perceptual loss to supervise the predicted residual $\Delta \hat{I}$, much like how it's applied to the final image:
\begin{equation}
\mathcal{L}_{\text{perc}}^{\text{res}} = \|\phi(\Delta I)) - \phi(\Delta \hat{I})\|_2^2
\end{equation}

While this may seem counterintuitive at first, our experiments demonstrate that this approach ensures successful training, as shown in \cref{tab:loss_table}. This could be because applying perceptual loss to the residual encourages the model to focus more on structural information, which enhances supervision of the sparse signal.


\begin{figure}[tb]
    \centering
    \includegraphics[width=0.8\linewidth]{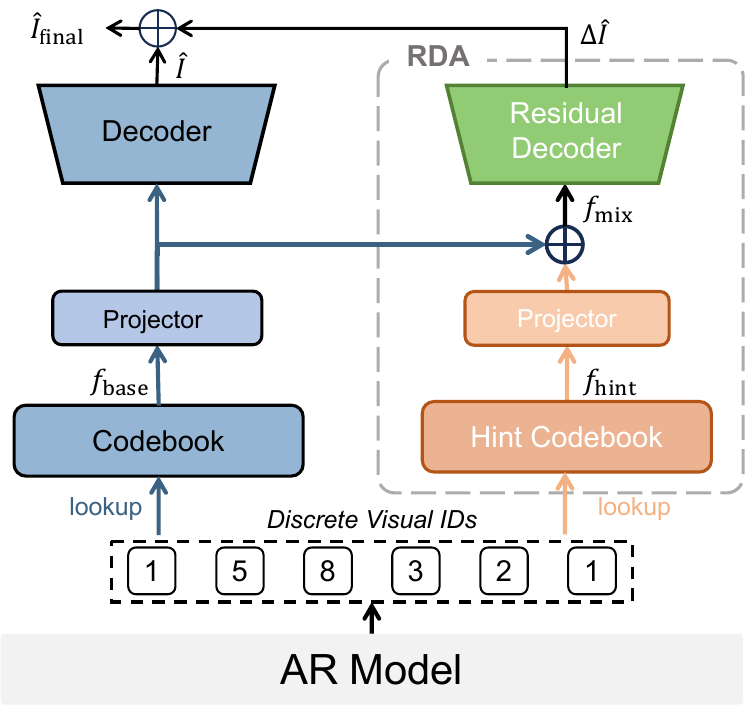}
    \caption{ \textbf{Inference pipeline of the AR model with equipped RDA.} RDA uses the same IDs to refine the generated image.
    } 
    \label{fig:ar_inference}
\end{figure}



%
\noindent\textbf{Edge-aware Sobel loss.}
We use the Sobel edge mask $M_{\text{edge}}$~\cite{ledig2017photorealisticsingleimagesuperresolution,wang2022edgeenhancedfeaturedistillationnetwork} to emphasize residuals along strong gradients and preserve edge structures at the pixel level, which can be formulated as \; $\mathcal{L}_{\text{sobel}} = |M_{\text{edge}} \odot (\Delta I - \Delta \hat{I})|_1$.

\begin{figure*}[tb]
    \centering
    \includegraphics[width=1\linewidth]{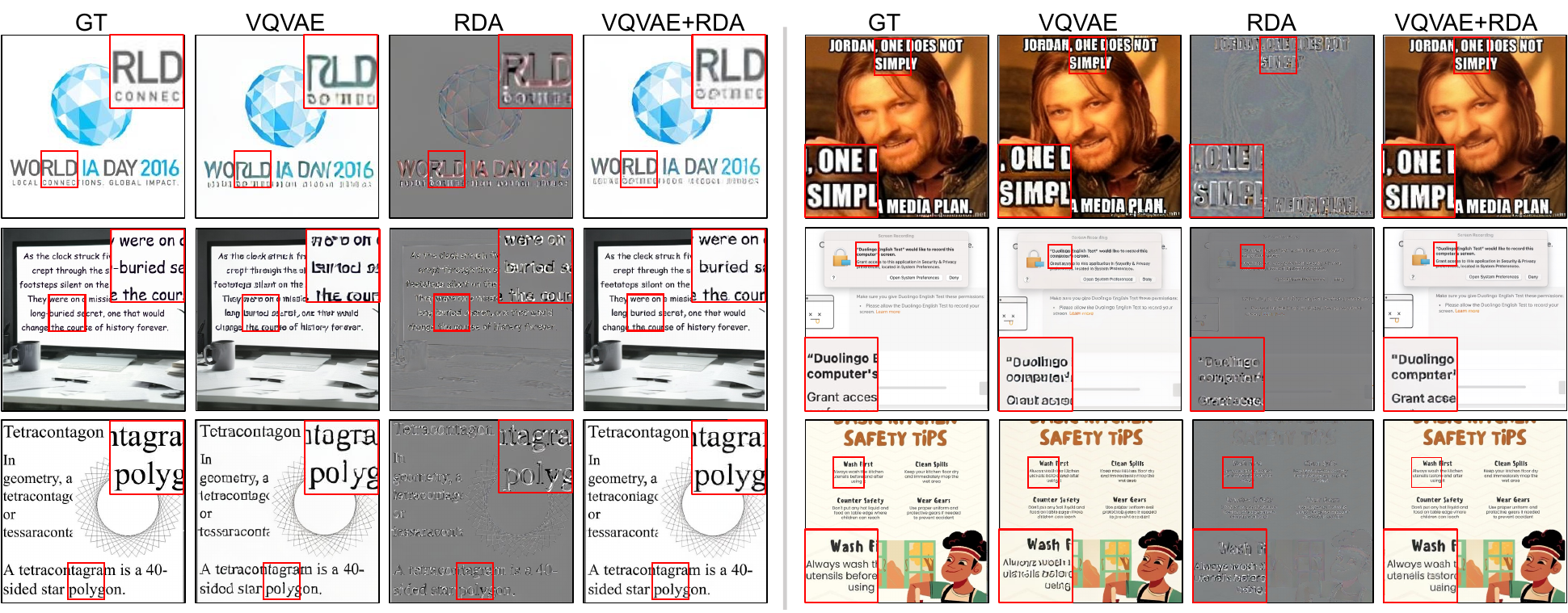}
    \caption{ 
    \textbf{Reconstruction performance of image tokenizer equipped with RDA.} We use LlamaGen-VQ as the VQVAE tokenizer.
    }
    \label{fig:tokenizer_recon}
\end{figure*}



%

\noindent\textbf{Frequency-domain loss.}
To preserve high-frequency details of residual image in the spectral domain:
\begin{equation}
\mathcal{L_{\text{freq}}} = \frac{1}{BCHW}\sum M_q \odot |\mathcal{F}_p - \mathcal{F}_t|^2
\end{equation}
where $\mathcal{F}_t = \mathrm{FFT}(\Delta I)$, $\mathcal{F}_p = \mathrm{FFT}(\Delta \hat{I})$, $\mathrm{FFT}$ denotes Fast Fourier Transforms and $M_q$ is a high-pass mask.
 
The final loss integrates all components:

\begin{equation}
\mathcal{L}_{\text{total}} =
 \mathcal{L}_{\text{rec}}
+ \mathcal{L}_{\text{perc}}^{\text{final}}
+ \mathcal{L}_{\text{perc}}^{\text{res}}
+ \mathcal{L}_{\text{sobel}}
+ \mathcal{L}_{\text{freq}}.
\end{equation}

During training, we freeze the base tokenizer and optimize only the RDA module.

\subsection{Plug-and-Play Inference in AR model}
During inference time, an AR model predicts a sequence of discrete indices $\{ i_n | i_n \in \{1, 2, \dots, K\} \}_{n=1}^{N}$. We can use those indices to reconstruct the initial coarse image $\hat{I}$ with the frozen decoder  $\mathcal{D}$.

Our hint codebook $Z'$ shares the same index space with the original codebook $Z$ through the Shared-ID mechanism.
For each token index, we retrieve the corresponding embeddings from both codebooks to obtain the base feature $f_{\text{base}}$ and hint feature $f_{\text{hint}}$, respectively.
These features are then passed through their projector and fused to form the representation $f_{\text{mix}}$.
The residual image $\Delta \hat{I}$ is then predicted by the residual decoder $\mathcal{D}_{\text{res}}$, yielding the refined output:
\begin{equation*}
\hat{I}_{\text{final}} = \hat{I} + \Delta \hat{I}.
\end{equation*}

\method can be seamlessly integrated into any AR model as a plug-and-play module at inference.

\begin{table*}[t]
\centering
\footnotesize
\caption{\textbf{Comparison of general AR models before and after applying \method.} Each cell shows the result  \dualcolor{\textit{\textbf{w}}}{\textit{\textbf{wo}}}  applying RDA. The Acc. and NED. on LongTextBench and CVTG-2K are multiplied by 100 for clarity.}
\resizebox{\textwidth}{!}{

\begin{tabular}{l|r|r|cc|cc|c|ccc}
\toprule
\multirow{2}{*}{\centering \textbf{Model}} &
\multirow{2}{*}{\centering \textbf{Size}} &
\multirow{2}{*}{\centering \textbf{Res}} &
\multicolumn{2}{c|}{\textbf{AnyText-Benchmark}} &
\multicolumn{2}{c|}{\textbf{Mario-Eval}} &
\multicolumn{1}{c|}{\textbf{LongTextBench}} &
\multicolumn{3}{c}{\textbf{CVTG-2K}} \\
& & & Acc.~$\uparrow$ & F1.~$\uparrow$ & Acc.~$\uparrow$ & F1.~$\uparrow$
& Acc.~$\uparrow$  & Acc.~$\uparrow$ & NED.~$\uparrow$ & ClipScore.~$\uparrow$ \\
\midrule

h

\multirow{2}{*}{\makecell[l]{Janus Pro}}
& 1B & 384  &
\dualcolor{2.05}{\textbf{2.56}} &
\dualcolor{2.76}{\textbf{3.21}} &
\dualcolor{1.75}{\textbf{2.06}} &
\dualcolor{2.37}{\textbf{2.57}} &
\dualcolor{0.14}{\textbf{3.89}} &
\dualcolor{0.18}{0.18} &
\dualcolor{13.55}{\textbf{15.06}} &
\dualcolor{\textbf{0.52}}{0.51} \\

& 7B & 384  &
\dualcolor{8.85}{\textbf{10.07}} &
\dualcolor{11.23}{\textbf{11.92}} &
\dualcolor{6.75}{\textbf{8.33}} &
\dualcolor{8.70}{\textbf{9.76}} &
\dualcolor{0.47}{\textbf{0.96}} &
\dualcolor{1.65}{\textbf{2.01}} &
\dualcolor{20.14}{\textbf{22.59}} &
\dualcolor{\textbf{0.64}}{0.63} \\

\addlinespace[2pt]
\hline
\addlinespace[2pt]
\multirow{4}{*}{TAR}
& 1.5B & 512  &
\dualcolor{13.53}{\textbf{14.62}} &
\dualcolor{14.75}{\textbf{15.20}} &
\dualcolor{10.57}{\textbf{11.98}} &
\dualcolor{11.08}{\textbf{11.53}} &
\dualcolor{3.05}{\textbf{3.89}} &
\dualcolor{3.49}{\textbf{4.01}} &
\dualcolor{27.69}{\textbf{30.22}} &
\dualcolor{0.68}{0.68} \\
& 1.5B & 1024 &
\dualcolor{12.57}{\textbf{13.11}} &
\dualcolor{\textbf{12.45}}{12.34} &
\dualcolor{10.90}{\textbf{11.41}} &
\dualcolor{\textbf{9.99}}{9.69} &
\dualcolor{\textbf{4.55}}{4.20} &
\dualcolor{3.92}{\textbf{4.08}} &
\dualcolor{32.66}{\textbf{34.42}} &
\dualcolor{\textbf{0.69}}{0.68} \\
& 7B & 512 &
\dualcolor{30.92}{\textbf{32.43}} &
\dualcolor{32.46}{\textbf{32.84}} &
\dualcolor{25.46}{\textbf{27.89}} &
\dualcolor{25.92}{\textbf{26.66}} &
\dualcolor{6.92}{\textbf{7.22}} &
\dualcolor{16.35}{\textbf{19.22}} &
\dualcolor{47.63}{\textbf{52.00}} &
\dualcolor{0.76}{0.76} \\
& 7B & 1024 &
\dualcolor{28.10}{\textbf{28.47}} &
\dualcolor{27.98}{\textbf{28.11}} &
\dualcolor{24.29}{\textbf{24.51}} &
\dualcolor{\textbf{22.94}}{22.77} &
\dualcolor{\textbf{6.33}}{5.68} &
\dualcolor{16.84}{\textbf{17.04}} &
\dualcolor{52.76}{\textbf{53.20}} &
\dualcolor{0.75}{0.75} \\

\addlinespace[2pt]
\midrule
\rowcolor{grpgray}\multicolumn{11}{c}{{\textit{Chameleon Tokenizer}}} \\
\addlinespace[3pt]

\multirow{1}{*}{Lumina-mGPT}
& 7B & 512 &
\dualcolor{2.12}{\textbf{2.29}} &
\dualcolor{2.44}{\textbf{2.62}} &
\dualcolor{0.29}{\textbf{0.32}} &
\dualcolor{0.47}{\textbf{0.54}} &
\dualcolor{0.14}{\textbf{0.28}} &
\dualcolor{\textbf{0.12}}{0.10} &
\dualcolor{13.19}{\textbf{14.16}} &
\dualcolor{0.65}{0.65} \\
\bottomrule

\end{tabular}
}

\label{tab:main_ar_model_table}
\end{table*}



\begin{table*}[ht]
\centering

\footnotesize

\caption{\textbf{Comparison of text-specific AR models before and after applying \method}. RDA consistently improves accuracy, F1, and CER across all subsets and resolutions, demonstrating  gains on 1B and 7B AR models. Each cell shows the result \dualcolor{\textit{\textbf{w}}}{\textit{\textbf{wo}}} applying RDA.}
\resizebox{\textwidth}{!}{
\begin{tabular}{l|r|r|ccc|ccc|ccc}
\toprule
\multirow{2}{*}{\centering \textbf{Model}} & \multirow{2}{*}{\centering \textbf{Size}} &
\multirow{2}{*}{\centering \textbf{Res}} & \multicolumn{3}{c|}{\textbf{StyledTextVisionBlend}} & \multicolumn{3}{c|}{\textbf{StyledTextSynth}} & \multicolumn{3}{c}{\textbf{TextScenesHQ}} \\
 && & {Acc.$\uparrow$} & {F1.$\uparrow$} & {CER.$\downarrow$}
 & {Acc.$\uparrow$} & {F1.$\uparrow$} & {CER.$\downarrow$} & {Acc.$\uparrow$} & {F1.$\uparrow$} & {CER.$\downarrow$} \\
\midrule

Janus Pro* & 1B & 1024 &
\dualcolor{24.52}{\textbf{58.26}} &
\dualcolor{29.85}{\textbf{63.18}} &
\dualcolor{0.47}{\textbf{0.23}} &
\dualcolor{12.75}{\textbf{36.81}} &
\dualcolor{17.37}{\textbf{41.37}} &
\dualcolor{0.75}{\textbf{0.52}} &
\dualcolor{1.35}{\textbf{1.76}} &
\dualcolor{2.03}{\textbf{2.38}} &
\dualcolor{0.85}{\textbf{0.83}} \\

Lumina-mGPT* & 7B & 512 &
\dualcolor{56.31}{\textbf{69.78}} &
\dualcolor{60.22}{\textbf{74.02}} &
\dualcolor{0.26}{\textbf{0.18}} &
\dualcolor{2.56}{\textbf{5.14}} &
\dualcolor{3.51}{\textbf{6.83}} &
\dualcolor{0.80}{\textbf{0.77}} &
\dualcolor{2.72}{\textbf{3.59}} &
\dualcolor{3.62}{\textbf{5.07}} &
\dualcolor{0.83}{\textbf{0.82}} \\

Lumina-mGPT* & 7B & 1024 &
\dualcolor{60.50}{\textbf{73.14}} & 
\dualcolor{64.53}{\textbf{77.57}}& 
\dualcolor{0.25}{\textbf{0.17}} &
\dualcolor{34.33}{\textbf{48.96}} &
\dualcolor{37.91}{\textbf{53.43}} &
\dualcolor{0.42}{\textbf{0.32}} &
\dualcolor{9.78}{\textbf{11.01}} &
\dualcolor{11.16}{\textbf{12.07}} &
\dualcolor{0.79}{\textbf{0.78}} \\

\bottomrule
\end{tabular}
}
\label{tab:gen_text_atlas}
\end{table*}



\section{Experiments}
\label{sec:experiments}

\subsection{Implementation details}

We build the RDA on LlamaGen-VQ using \textit{vq\_ds16\_t2i}~\cite{llamagen} and on Chameleon-VQ~\cite{chameleon} using its \textit{built-in VQ-VAE}. The Hint codebook sizes are 16,384 and 8,192, respectively, both with dimension 16. We train our RDA from scratch while keeping the base tokenizer frozen. Training is conducted at $256{\times}256$ resolution for 120k steps with a global batch size of 512 on 64 V100 GPUs (8 nodes $\times$ 8 GPUs), using mixed precision, AdamW~\cite{adamw} ($\beta_1{=}0.9$, $\beta_2{=}0.95$) with learning rate 1e-4, and gradient clipping at 1. RDA is trained on Mario-10M~\cite{chen2023textdiffuserdiffusionmodelstext}. From the training split, we exclude all samples overlapping the AnyText-Benchmark~\cite{tuo2024anytextmultilingualvisualtext}, then randomly sample 5M images.

\subsection{Evaluation Protocol}
We assess RDA in two settings: (i)AR model generation: improving text-to-image generation for AR models; (ii) Text Image Reconstruction: enhancing tokenizer reconstruction fidelity on text-centric data.

\noindent \textbf{Generation.} We adapt three AR models: Janus Pro and TAR (with an RDA-adapted LlamaGen-VQ), and Lumina-mGPT~\cite{lumina_mgpt} (with an RDA-adapted Chameleon-VQ). Models are evaluated on five text-oriented datasets: Mario-Eval~\cite{chen2023textdiffuserdiffusionmodelstext}, AnyText-Benchmark~\cite{tuo2024anytextmultilingualvisualtext}, LongTextBench~\cite{longtextbench}, CVTG-2K~\cite{du2025textcrafteraccuratelyrenderingmultiple}, and TextAtlasEval~\cite{wang2025textatlas5mlargescaledatasetdense}. For Mario-Eval and AnyText-Benchmark, we report text-specific metrics—OCR Accuracy(Acc.) and OCR F1(F1.)—computed with PaddleOCR~\cite{paddleocr_vl,paddle_ocr}.For LongTextBench, CVTG-2K, and TextAtlasEval, we follow the official eval protocols.

\noindent \textbf{Reconstruction.} We evaluate RDA-adapted visual tokenizers on three text-centric sets: AnyText-Benchmark~\cite{tuo2024anytextmultilingualvisualtext}, Mario-Eval~\cite{chen2023textdiffuserdiffusionmodelstext}, and the TextAtlasEval-StyledTextSynth subset~\cite{wang2025textatlas5mlargescaledatasetdense}. AnyText-Benchmark and Mario-Eval are tested at 256 and 512 image size; StyledTextSynth at 512 and 1024 image size. We report OCR Accuracy/F1, SSIM and LPIPS. Comparisons include six models: LlamaGen-VQ, its RDA-adapted version, Chameleon-VQ, its RDA-adapted version, TA-Tok~\cite{tar}, and UniTok~\cite{unitok}.

\begin{table*}[t]
\centering
\footnotesize
\caption{\textbf{Comparison of text image reconstruction performance across image tokenizers.} AR-Free means that the proposed module can be attached to the AR model without any retraining. AnyText-Benchmark and Mario-Eval are evaluated at $256$ and $512$ , while  StyledTextSynth is evaluated at $512$ and $1024$. SSIM and LPIPS are multiplied by 100 for clarity.}
\resizebox{\linewidth}{!}{%
\begin{tabular}{l|c|c|c c c c|cccc|cccc}
\toprule
\multirow{2}{*}{\centering \textbf{Model}} &
\multirow{2}{*}{\centering \textbf{Data}} &
\multirow{2}{*}{\centering 
\makecell[c]{\textbf{AR} \\ \textbf{Free}}} &
\multicolumn{4}{c|}{\textbf{AnyText-Benchmark}} &
\multicolumn{4}{c|}{\textbf{Mario-Eval}} &
\multicolumn{4}{c}{\textbf{StyledTextSynth}} \\
& & & Acc. & F1. & SSIM~ & LPIPS 
& Acc. & F1. & SSIM & LPIPS  & Acc. & F1. & SSIM & LPIPS\\
\midrule

\rowcolor{grpgray}\multicolumn{15}{c}{{\textit{Low Resolution}}} \\
\addlinespace[3pt]

LlamagenVQ & 50M &  \ding{55}  & 21.26 & 28.34  & 71.66 & 19.12  &15.26 & 21.31 & 69.82 & 18.82 &4.12	&6.55&	66.33	&22.39	\\
\rowcolor{myBlue!60}   \textit{w/ \method} & 5M & \ding{52}  & 36.79 & 44.20 & 74.11 & 18.32  & 27.59	& 34.86&	69.79&	19.00  & 13.57	&18.18&	67.67	&22.23	\\

ChameleonVQ & - &\ding{55}  &15.97&20.52  & 69.85&20.98  &11.05&14.92 & 65.51&21.94 &5.52&7.44&	63.81&26.89	\\
\rowcolor{myBlue!60}   \textit{w/ \method} & 5M & \ding{52}  & 27.51&34.03 & 71.89&20.09  & 19.58&25.68&	67.32&21.09  & 7.80&10.53&	65.13	& 24.90	\\

\addlinespace[2pt]
\hline
\addlinespace[2pt]
TA-Tok & 200M & \ding{55}   & -& - & - & - & - & - & - & - &2.25&3.93&44.35	&47.73\\
UniTok & 1.28B & \ding{55}    & 62.61 &24.49&79.83&14.04 & 55.71	&61.30 &82.60&14.11&62.11&	68.54&92.43&11.32 \\

\addlinespace[2pt]
\hline
\rowcolor{grpgray}\multicolumn{15}{c}{{\textit{High Resolution}}} \\

\addlinespace[2pt]

LlamagenVQ & 50M & \ding{55}  & 58.04 & 64.55 & 83.11 & 6.52  & 47.52 &	55.30 & 79.67	& 7.87	&55.58	&60.51	&75.58&	8.95	\\
\rowcolor{myBlue!60} \textit{w/ \method} & 5M& \ding{52}  & 66.48 & 70.95 & 84.55 & 6.28  &57.36& 25.16 &	81.50&	7.87	 &69.49	&74.22&	77.03	&9.70	\\

ChameleonVQ & - &\ding{55}  & 53.91&59.17 & 81.90	&7.04  & 42.33&48.05& 78.03	& 8.67	&39.01& 43.82	&71.63	&10.86	\\
\rowcolor{myBlue!60} \textit{w/ \method} & 5M & \ding{52}  & 61.28	& 66.26 & 82.75	& 6.77  &50.42&55.42 &	79.09	&8.46	 &73.74	& 9.99&	77.03	&9.70	\\

\addlinespace[2pt]
\hline
\addlinespace[2pt]

TA-Tok & 200M & \ding{55}   & 40.58&45.94 & 57.58	& 25.21 & 32.48&	37.55 & 52.91&29.24&6.69&9.24&42.70	&46.45\\
UniTok & 1.28B & \ding{55}   & 74.60&74.50 &91.16&3.12&80.20&82.18&79.83&14.04&84.92&87.69&99.73&00.01\\

\bottomrule

\end{tabular}
}

\label{tab:tokenizer_performance}
\end{table*}





\subsection{Main Results.}

\noindent\textbf{Generation tasks}
As shown in \cref{tab:main_ar_model_table}, RDA integrates into \textit{general AR models} without retraining and consistently boosts performance. When applied to Janus Pro 7B, it improves Acc. from 8.85\%→10.07\% on AnyText-Benchmark and from 6.75\%→8.33\% on Mario-Eval. Since our method operates at the visual tokenizer level, a single model trained with the LlamaGen tokenizer transfers plug-and-play to both Janus Pro and TAR (across model sizes), yielding consistent gains in text rendering accuracy regardless of AR architecture or scale. Trained only at \textit{256}px, RDA also generalizes to higher resolutions (384, 512, 1024px) without fine-tuning. Those experiments show that our method is \textit{an efficient universal adapter that enhances text generation quality with negligible computational overhead}.

 We also test our method on \textit{AR models fine-tuned on text-rich TextAtlas dataset}. As shown in \cref{tab:gen_text_atlas}, RDA yields substantially larger improvements on these text-specialized models. For Janus-Pro 1B on StyledTextVisionBlend, OCR accuracy rises from 24.52\% to 58.26\% (+33.74 points) and F1 from 29.85\% to 63.18\% (+33.33 points). This improvement is remarkably larger than on general models: RDA boosts the general Janus-Pro 7B by only +1.22 points (8.85\%→10.07\%) on AnyText-Benchmark, but enhances the text-tuned Janus-Pro 1B by +33.74 points (24.52\%→58.26\%) on StyledTextVisionBlend.

This disparity reveals that general AR models face a dual bottleneck: weak text token prediction and limited reconstruction fidelity. Text-specific fine-tuning addresses the former, making the tokenizer decoder the dominant bottleneck. Consequently, RDA directly addresses this bottleneck: by enhancing tokenizer reconstruction without modifying the AR model, it fully realizes the improved token predictions, yielding more accurate text rendering.

\noindent\textbf{Reconstruction tasks}
As shown in \cref{tab:tokenizer_performance}, equipping image tokenizer with RDA consistently improves all metrics—most notably the text-specific ones. OCR accuracy and F1 rise by around 10 points, and SSIM/LPIPS also improve, validating the effectiveness of our residual refinement framework. Compared with recent visual tokenizers, RDA is (i) \textit{data-efficient}: trained on just a small subset 5M instead of hundreds of millions to billions of images (e.g., UniTok uses 1.28B images~\cite{unitok}); and (ii) \textit{AR-training free}: it preserves the original token ID space, avoiding shifts in token distributions and eliminating the need to retrain downstream AR models.
In contrast, other methods inevitably alter the token distribution, necessitating complete retraining of all downstream AR models—a prohibitively expensive process that can take thousands of GPU hours for billion-parameter systems.

\subsection{Ablation Study}

\noindent\textbf{Ablation on Loss Functions.} We perform an ablation study on each loss function in \cref{tab:loss_table}. The results highlight that the perceptual loss on the residual $\mathcal{L}_{\text{perc}}^{\text{res}}$  
is the most critical, boosting baseline OCR accuray from 58.04\% to 66.48\%. Removing this loss leads to a failure in residual training, with performance degenerating to the baseline. This emphasizes that perceptual supervision on residuals is crucial for optimizing sparse residual signals, as it encourages the model to focus more on structural information, which enhances the supervision of the sparse signal. The $\mathcal{L}_{\text{sobel}}$  is also important for text rendering, improving OCR accuracy by 2.17\% through edge-aware guidance. Both $\mathcal{L}_{\text{perc}}^\text{final}$ and $L_{\text{freq}}$ contribute marginal but consistent improvements.

\begin{table}[tb]
\centering
\small
\caption{\textbf{Ablation on loss composition.} Residual perceptual loss$\mathcal{L}_{\text{perc}}^{\text{res}}$ is critical—removing it causes complete training failure.}
\resizebox{\linewidth}{!}{
\begin{tabular}{l|cc|ccc}
\toprule
 &Acc. & F1. & PSNR & SSIM &  LPIPS \\
\midrule
Baseline &58.04 &64.55 &24.33&	83.11&	6.52 \\
\rowcolor{myBlue!60} Full (ours)           &\textbf{66.48}	&\textbf{70.95}  & \textbf{25.73}	&84.55&	6.28	\\
w/o $\mathcal{L}_{\text{freq}}$ &66.13	&70.36   & 25.70&	\textbf{84.57}&	\textbf{6.24}	\\
w/o $\mathcal{L}_{\text{sobel}}$ &64.31	&69.39 & 25.47&	84.07&	6.44	\\
\rowcolor{grpgray} w/o  $\mathcal{L}_{\text{perc}}^{\text{res}}$ &58.30	&64.84& 24.46&	83.06	&6.49	 \\
w/o $\mathcal{L}_{\text{perc}}^{\text{final}}$ &64.58&	69.87&25.59&	83.95&	7.63	 \\
\bottomrule
\end{tabular}
}
\label{tab:loss_table}
\end{table}


\noindent\textbf{Effect on Hint Codebook design and Residual learning paradigm.} \cref{tab:codebook_table} ablates the hint codebook design.
In the “No Codebook” setting, the residual decoder $\mathcal{D}_{res}$ only receives frozen base features $f{base}$ from the original codebook without any hint embeddings.
Even under this condition, residual learning alone improves accuracy from 58.04\% to 64.25\%.
Adding the hint codebook brings consistent gains across dimensions (8, 16, 32), with dim=16 achieving the best performance (66.48\%).
Notably, training \textit{with} the hint codebook but testing \textit{without} it ($16^\dagger$: 65.45\%) still outperforms “No Codebook” (64.25\%), suggesting that the hint codebook enhances residual representation learning during training beyond its direct contribution at inference.

\begin{figure*}[t]
    \centering
    \includegraphics[width=1\linewidth]{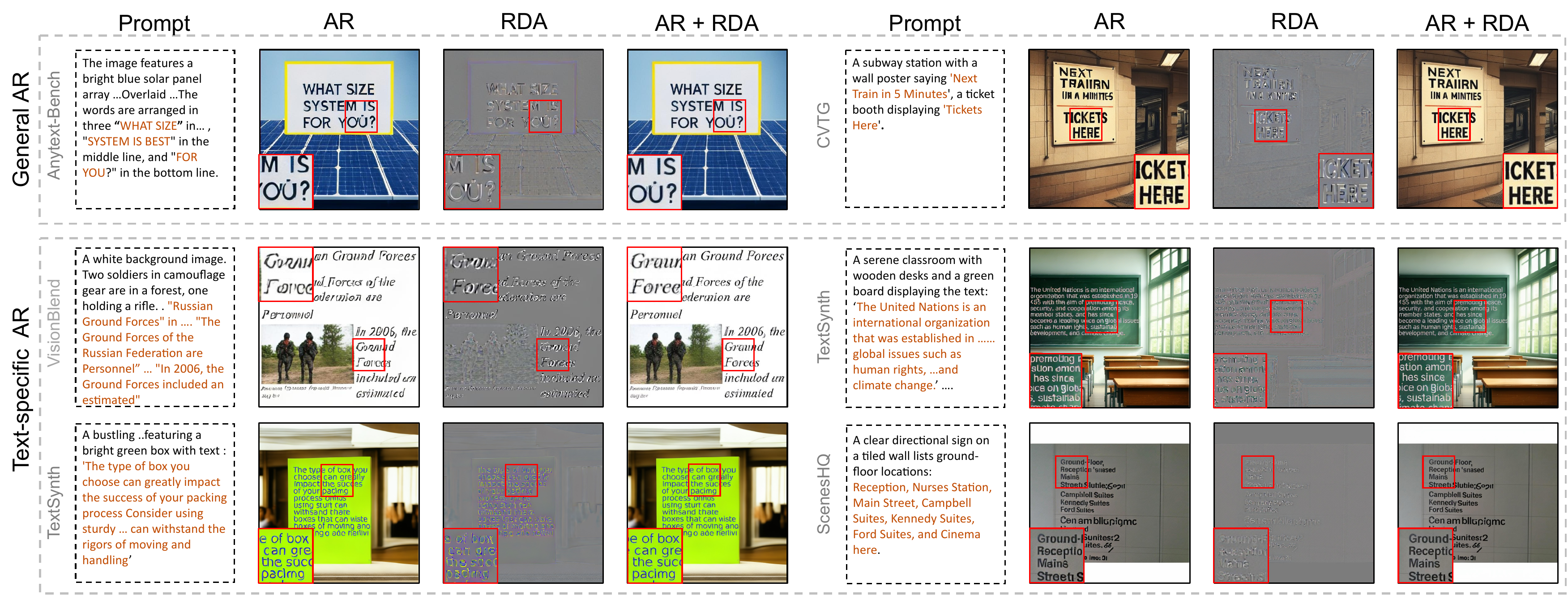}
    \caption{ 
    \textbf{Qualitative generation results of AR models.}
    General AR uses TAR-7B at 512 resolution, while text-specific AR uses Janus-Pro$^*$ 1B 1024 (left) and Lumina-mGPT$^*$ 7B 1024 (right). Please zoom in for better visualization of text details. More visual results are provided in the Appendix.
    } 
    \label{fig:ar_gen}
\end{figure*}



%

\noindent\textbf{Importance of Feature Injection.} \cref{tab:feat_add_table} shows the importance of injecting the base feature $f_{\text{base}}$ into the residual decoder. In the No-Injection setting, only the hint codebook feature $f_{\text{hint}}$ is used during training. Despite retaining the hint codebook, this configuration performs even worse than the baseline, indicating that instance-dependent feature injection is crucial for residual decoder to effectively leverage the hint codebook and learn meaningful refinements. Among fusion methods (Add, Cross-Attention, Concat+Conv), performance is comparable; we use simple addition for efficiency.
\begin{table}[tb]
\centering
\small
\caption{ \textbf{Ablation on hint codebook design and resiual learning paradigm}. $^{\dagger}$ notes train with hint codebook but tests without it.}
\begin{tabular}{l|cc|ccc}
\toprule
 & Acc. & F1.& PSNR & SSIM & LPIPS \\
\midrule
Baseline &58.04 &64.55 &24.33&	83.11&	6.52 \\
\rowcolor{grpgray} No Codebook            & 64.25 & 68.72& 25.48 & 84.12&6.44 \\
dim=8  & 64.97	& 69.65& 25.56&	84.27 &6.36\\
dim=16$^{\dagger}$  & 65.45 &70.02&25.48  &84.35 &6.38  \\
\rowcolor{myBlue!60} dim=16  & \textbf{66.48} &\textbf{70.95} &\textbf{25.73}  &\textbf{84.55}& \textbf{6.28}  \\
dim=32 & 65.73 & 70.24 & 25.57 &  84.29&6.36\\
\bottomrule
\end{tabular}
\label{tab:codebook_table}
\end{table}


\begin{table}[t]
\centering
\footnotesize
\caption{\textbf{Ablation on instance-dependent feature fusion strategy}. Code.: hint codebook; Fuse.: feature injection.}
\setlength{\tabcolsep}{2.5pt}\renewcommand{\arraystretch}{1.05}
\begin{tabular}{l|c|c|cc|ccc}
\toprule
 &Code. &Fuse.& Acc$\uparrow$ & F1$\uparrow$& PSNR$\uparrow$ & SSIM$\uparrow$ & LPIPS$\downarrow$\\
\midrule
Baseline &\ding{55} &\ding{55} &58.04 &64.55 &24.33&	83.11&	6.52 \\
\rowcolor{grpgray} No Injection  &\ding{52} &\ding{55} & 58.04 & 64.55& 24.21 & 83.08 &6.62\\
\rowcolor{myBlue!60} Add (ours)  &\ding{52} &\ding{52} & \textbf{66.48} &\textbf{70.95}&\textbf{25.73}  &\textbf{84.55}& 6.28  \\
Cross-Attn.  &\ding{52} &\ding{52} & 65.58&	70.13  &\textbf{25.73}&	84.46 &\textbf{6.27}  \\
Concat + Conv  &\ding{52} &\ding{52} &66.04&70.86 &25.70&84.47&6.31  \\

\bottomrule
\end{tabular}

\label{tab:feat_add_table}
\end{table}


\subsection{Discussion}

\begin{figure}
  \centering
\includegraphics[width=1\linewidth]{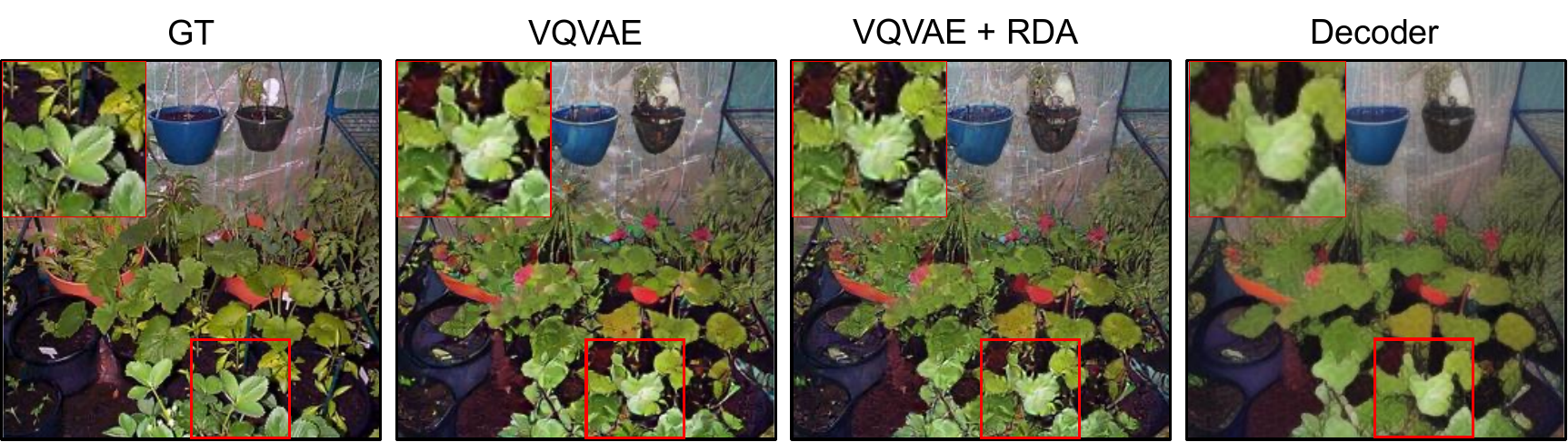}
  \caption{\textbf{Comparison of RDA with direct decoder fine-tuning.}
 RDA preserves fine-grained details and sharper textures, while direct decoder fine-tuning leads to over-smoothing.}
  \label{fig:decoder}
\end{figure}

\noindent\textbf{Out-of-Distribution Generalization.}
A natural alternative is to fine-tune the decoder on text-rich data. However, this approach suffers from poor out-of-distribution (OOD) generalization. 
We evaluate on ImageNet-10K validation set—a natural image-dominated benchmark unseen during training (models trained on text-rich Mario-10M). 
\cref{tab:imagenet_generalization} shows that while fine-tuning improves in-distribution pixel metrics, it catastrophically degrades OOD quality: FID worsens from 9.63 to 14.75 (+5.12) and LPIPS from 0.5144 to 0.5473, with visible over-smoothing (\cref{fig:decoder}).
In contrast, RDA maintains OOD robustness (FID: 11.20, +1.57) by freezing the base decoder and learning only additive residual corrections, preserving general image priors while enhancing text quality.

\begin{table}[t]
\centering
\caption{\textbf{Out-of-distribution generalization on ImageNet validation set.} Models are trained on text-rich Mario-10M but tested on natural image-dominated ImageNet.}
\begin{tabular}{l|cccc}
\toprule
 & FID$\downarrow$ & PSNR$\uparrow$ & SSIM$\uparrow$ & LPIPS$\downarrow$ \\
\midrule
Base   & 9.63  & 14.58 & 34.97 & 51.44 \\
Decoder    & 14.75 & 15.01 & 37.96 & 54.73 \\
\rowcolor{myBlue!60} Ours  & 11.20 & 14.87 & 35.90 & 50.95 \\
\bottomrule
\end{tabular}
\label{tab:imagenet_generalization}
\end{table}

%

\noindent\textbf{Difference between RQ-VAE.} 
While both our method and RQ-VAE contain the idea of residual learning, RDA \textit{differs fundamentally} in both purpose and functionality. \textit{In terms of purpose}, RQ-VAE uses residual learning to minimize the difference between quantized features and the original encoded features, whereas RDA directly learns the difference between the reconstructed image and the ground truth at the pixel level. \textit{In terms of functionally}, RQ-VAE alters the token distribution after training, requiring retraining of downstream AR models. In contrast, RDA operates without modifying the token space and uses the explicit Shared-ID strategy during training, making it compatible with pretrained AR models. Furthermore, our method is \textit{orthogonality to methods that improve the representation of the tokenizer, our method can also be readily applied to RQ-VAE}.

\subsection{Qualitative Results}
We present qualitative comparisons to illustrate the effectiveness of RDA in both generation and reconstruction settings.
\cref{fig:ar_gen} shows text-to-image generation results of AR models before and after applying it, where our method produces sharper and more legible text.
\cref{fig:tokenizer_recon} further visualizes reconstruction results of LlamaGen-VQ with and without RDA, demonstrating that RDA significantly enhances the text reconstruction fidelity across multiple resolutions.

\section{Conclusion}
\label{sec:conclusion}

We propose the Residual Decoder Adapter (RDA), which boosts text rendering in autoregressive models without retraining the tokenizer or AR model. By refining the tokenizer reconstruction with a paired hint codebook design and residual learning paradigm, RDA significantly improves text rendering performance on various competitive benchmarks. Importantly, it offers an efficient, non-invasive solution for enhancing text rendering while maintaining compatibility with pretrained models.

{
    \small
    \bibliographystyle{utils/ieeenat_fullname}
    \bibliography{main}
}

\maketitlesupplementary

\tableofcontents

\setcounter{page}{1}

\section{Implementation Details}
\subsection{Model Architecture}
\noindent\textbf{Hint Codebook}
We instantiate a paired codebook that mirrors the size and index space of the original tokenizer codebook. 
The embedding vectors are learned from scratch, but their indices remain aligned with the base codebook.

\noindent\textbf{Projector Design}
We use a lightweight $1{\times}1$ Conv2d to map hint-codebook embeddings ($d_{\text{hint}}=16$) 
to the residual decoder input space:
\begin{equation}
\text{Projector}_q: \text{Conv2d}(16, 256, 1)
\end{equation}

\noindent\textbf{Residual Decoder Architecture}
The architecture largely follows decoder of LlamaGenVQ~\cite{llamagen} design with key modifications for 
high-resolution text detail capture. The full specification is shown in ~\cref{tab:rd_arch_compact}.


\begin{table}[t]
    \centering
    \caption{\textbf{Residual Decoder architecture.} The decoder follows a simple expand–process–upsample design.}
    \label{tab:rd_arch_compact}
    \resizebox{0.95\linewidth}{!}{
    \begin{tabular}{l l c c}
        \toprule
        \textbf{Stage} & \textbf{Layer} 
        & \textbf{Input → Output} 
        & \textbf{Resolution} \\
        \midrule
        
        Input  & -         & [B, 256, h, w]   & 1/16 \\
        \midrule
        
        Expansion & Conv-in & 256 → 1024 & 1/16 \\
        \midrule

        \multirow{3}{*}{Processing}
            & ResBlock  & 1024 → 1024 & 1/16 \\
            & AttnBlock & 1024 → 1024 & 1/16 \\
            & ResBlock  & 1024 → 1024 & 1/16 \\
        \midrule

        \multirow{5}{*}{Upsampling}
            & Block 0 & 1024 → 512 & 1/16 → 1/8 \\
            & Block 1 & 512 → 512 & 1/8 → 1/4 \\
            & Block 2 & 512 → 512 & 1/4 → 1/2 \\
            & Block 3& 512 → 256 & 1/2 → 1/1 \\
            & Block 4 & 256 → 128 & 1/1 → 1/1 \\
        \midrule
        
        Output & Conv-out & 128 → 3 & 1/1 \\
        \bottomrule
    \end{tabular}}
\end{table}

\subsection{Training Configuration}

\begin{figure}[t]
  \centering
  \includegraphics[width=0.48\textwidth]{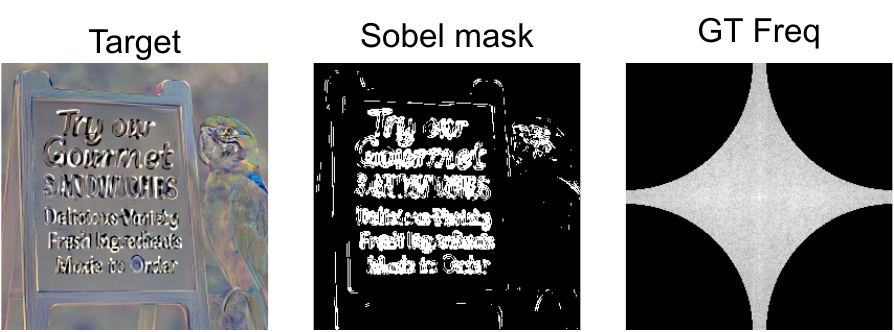} 
    \caption{\textbf{Visualization of Sobel mask and  frequency mask.} }

  \label{fig:sup_sobel_mask}
\end{figure}

All loss weights are set to 1.0. We use a high-pass mask $M_q$ with $q=0.8$ during training.
The gradient-sensitive loss $\mathcal{L}_{\text{sobel}}$ emphasizes edge structures, 
while $\mathcal{L}_{\text{freq}}$ preserves high-frequency components in Fourier space. 
Visualization of these masks is provided in \cref{fig:sup_sobel_mask}.

\section{Computational Cost Analysis}

\subsection{Parameters and Latency}
~\cref{tab:rda_cost} quantifies the computational overhead introduced by RDA 
on different AR models and tokenizers. All measurements are conducted in inference mode 
on a single V100 GPU.

RDA introduces negligible overhead ($<2\%$ latency), 
while achieving substantial improvements in text rendering quality.

\begin{table}[t]
\centering
\small
\caption{\textbf{Computational cost of RDA.} 
We report the overhead on both the tokenizer side and the end-to-end AR model.}
\label{tab:rda_cost}
\begin{tabular}{lcc}
\toprule
\textbf{Module} & \textbf{Params} & \textbf{Latency} \\
\midrule
\rowcolor{grpgray}\multicolumn{3}{c}{\textit{Tokenizer}} \\
LLamaGenVQ        & 72 M & 34.50 ms \\
 + RDA       & 237 M& 80.11 ms \\

ChameleonVQ        & 69 M & 34.48 ms \\
 + RDA       & 234 M & 80.90 ms \\

\midrule
\rowcolor{grpgray}\multicolumn{3}{c}{\textit{AR model}} \\
Janus Pro 1B         & 2.09 G  & 11.20 s \\
+ RDA  & 2.15 G (+7.89\%) & 11.36 s (+1.43\%)  \\
Janus Pro 7B         & 7.42 G & 14.63 s  \\
 + RDA  & 7.58 G (+2.22\%) & 14.80 s (+1.16\%) \\
Tar 7B         & 9.40G & 72.93 s  \\
+ RDA & 9.56 G (+1.76\%)  & 73.25 s (+0.42\%)\\
Lumina-mgpt         & 7.04 G & 212.16 s \\
 + RDA & 7.20 G (+2.34\%) & 212.43 s (+0.13\%)  \\
\bottomrule
\end{tabular}
\end{table}

\section{Understanding the Bottleneck}

\subsection{Tokenizer Reconstruction Limit}
\label{subsec:tokenizer_bottleneck}

\begin{figure}[t]
  \centering
  \includegraphics[width=0.48\textwidth]{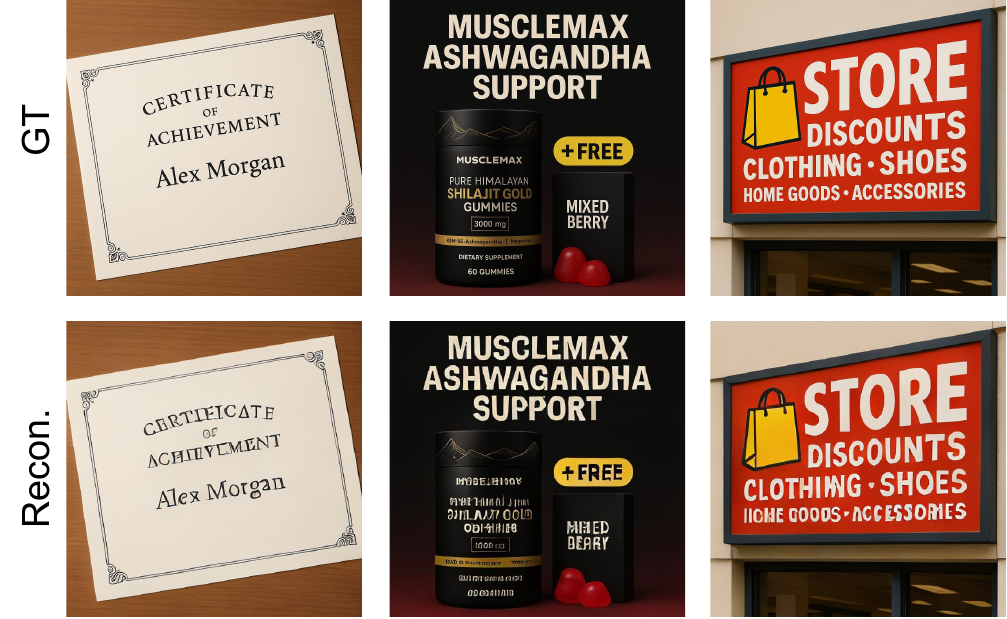} 
  \caption{\textbf{Comparison between ground-truth images and reconstructions from the base image tokenizer}. The reconstructed results blur fine-grained text strokes and distort glyph edges, indicating that textual details are significantly degraded during quantization and decoding.}
  \label{fig: sup tokenizer recon}
\end{figure}

We verify that the base tokenizer exhibits inherent reconstruction limitations 
even when provided perfect ground-truth input. ~\cref{fig: sup tokenizer recon} 
shows that the quantization-decoding process inherently loses fine-grained details, 
blurring text strokes and distorting glyph edges. This confirms that the tokenizer's 
reconstruction capability is the primary bottleneck.

\subsection{Dual Bottleneck in Text Rendering}

General AR models face two bottlenecks:
\begin{enumerate}
    \item \textbf{Token prediction:} Weak text token prediction from the AR model.
    \item \textbf{Reconstruction:} Limited reconstruction fidelity from the tokenizer.
\end{enumerate}

Text-specific fine-tuning addresses the first bottleneck, making the tokenizer decoder 
the dominant limitation. RDA directly targets this by enhancing reconstruction without 
modifying the AR model, enabling large improvements on text-tuned models.

\section{Failure Case Analysis}

We categorize failures into two types based on their source:

\subsection{Tokenizer-Level Failures}

\begin{figure}[t]
  \centering
  \includegraphics[width=0.48\textwidth]{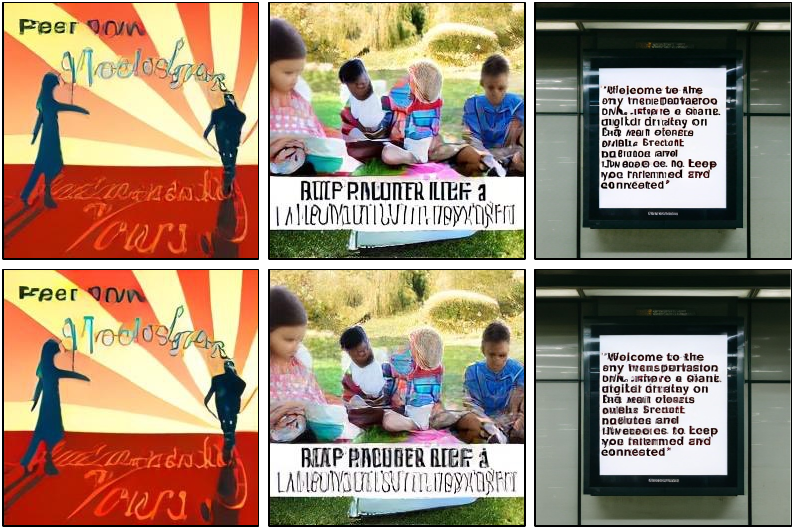} 
    \caption{\textbf{Failure Case of Tokenizer.} }

  \label{fig:sup_fail_case_tokenizer}
\end{figure}

When characters are extremely small or visually ambiguous, the tokenizer may assign 
incorrect visual tokens, leading to unrecoverable errors. ~\cref{fig:sup_fail_case_tokenizer} 
illustrates such cases.

\subsection{AR Model-Level Failures}

\begin{figure}[t]
  \centering
  \includegraphics[width=0.48\textwidth]{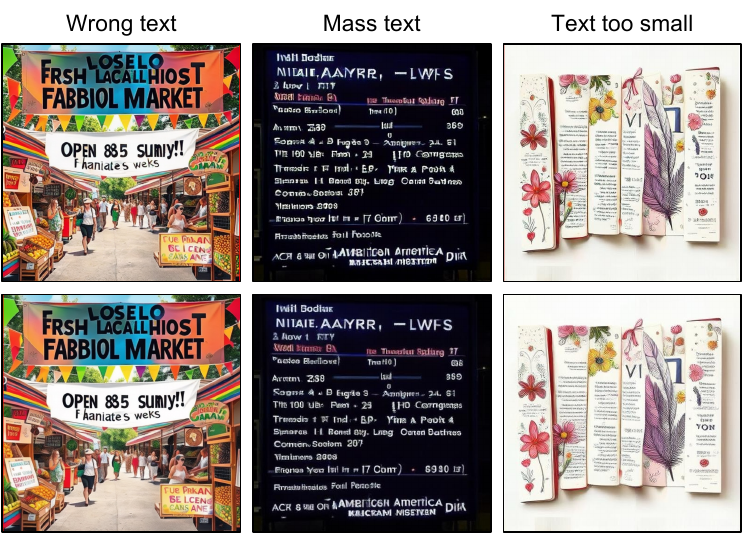} 
    \caption{\textbf{Failure Case of AR model.} }

  \label{fig:sup_fail_case_ar}
\end{figure}

When the AR model generates malformed glyph structures during generation, 
RDA cannot correct them since it operates on the decoded output. 
~\cref{fig:sup_fail_case_ar} shows examples where the AR model produces structurally 
incorrect characters.

\section{Training Analysis}

\subsection{Training Stability and Convergence}

\begin{figure}[t]
  \centering
  \includegraphics[width=0.48\textwidth]{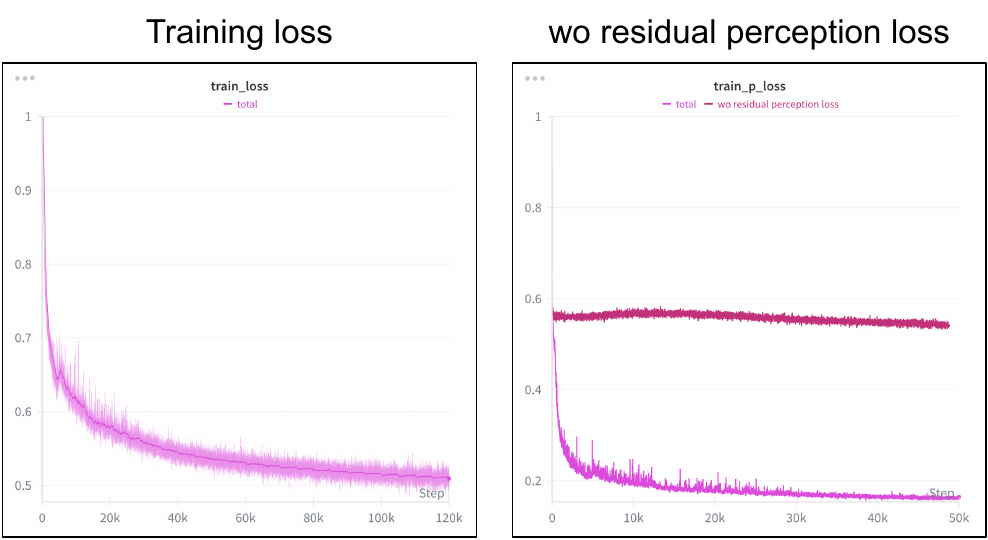} 
    \caption{\textbf{Visualization of loss curve.} }

  \label{fig:sup_loss_curve}
\end{figure}

~\cref{fig:sup_loss_curve} presents training curves for RDA. The optimization is stable throughout training. Critically, without residual perceptual loss 
$\mathcal{L}_{\text{perc}}^{\text{res}}$, the residual branch fails to converge and produces only blurry gray regions. This underscores the importance of perceptual supervision for learning meaningful high-frequency details.

\section{Design Justifications}

\subsection{Why Shared-ID Preserves Compatibility}
The Shared-ID mechanism ensures that the token ID distribution remains identical 
to the base tokenizer.

Since AR models learn a distribution over token IDs (not codebook embeddings), 
they can directly benefit from improved reconstruction without retraining.

\subsection{Why TAR Can Use RDA (LlamaGen-VQ)}

\begin{figure}[t]
  \centering
  \includegraphics[width=0.48\textwidth]{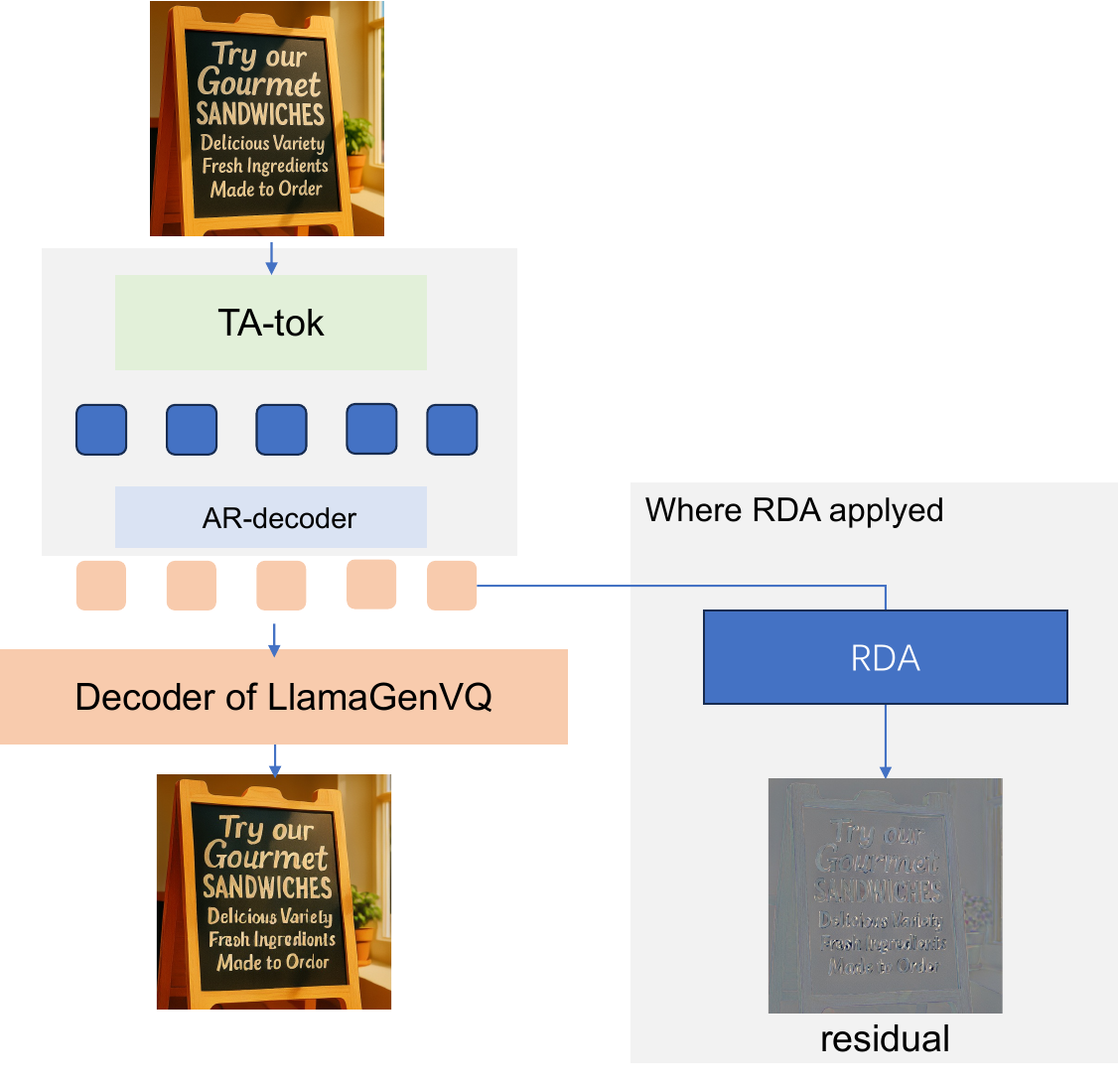} 
    \caption{\textbf{Where RDA is applied in TAR.} TAR modifies the tokenization and embedding pipeline before decoding, while RDA attaches after the decoder and refines pixel-level outputs without altering token IDs.}
    
  \label{fig:sup_tar_use}
\end{figure}

TAR adapts LlamaGen-VQ by modifying the tokenization and embedding pipeline before decoding. In contrast, RDA operates after the decoder and refines pixel-level outputs without altering token IDs. As illustrated in ~\cref{fig:sup_tar_use}, this separation allows RDA trained on LlamaGen-VQ to be directly applied to TAR without additional training.

\section{Extended Experiments}
\subsection{Tokenizer Results on Additional Datasets}
We also conduct evaluations on StyledTextVisionBlend and TextScenesHQ  of TextAtlasEval~\cite{wang2025textatlas5mlargescaledatasetdense, zhang2022distilling, zhang2025performing, zhang2025visual, zhang2024diff} to assess robustness. The results are summarized in ~\cref{tab:sub_tokenzier_add_tokenizer}.

\begin{table*}[t]
\centering

\caption{\textbf{Comparison of text image reconstruction performance across image tokenizers.}  StyledTextSynth is evaluated at $512$ and $1024$.}
\footnotesize
\begin{tabular}{l|c|c|c c c c|cccc}
\toprule
\multirow{2}{*}{\centering \textbf{Model}} &
\multirow{2}{*}{\centering \textbf{Data}} &
\multirow{2}{*}{\centering 
\makecell[c]{\textbf{AR} \\ \textbf{Free}}} &
\multicolumn{4}{c|}{\textbf{StyledTextVisionBlend}} &
\multicolumn{4}{c}{\textbf{TextScenesHQ}}  \\
& & & Acc. & F1. & SSIM~ & LPIPS 
& Acc. & F1. & SSIM & LPIPS \\
\midrule

\rowcolor{grpgray}\multicolumn{11}{c}{{\textit{Low Resolution}}} \\
\addlinespace[3pt]

LlamagenVQ & 50M &  \ding{55}  & 65.58& 71.34  & 82.80&	7.61  &11.11&16.35 & 50.36	& 46.27	\\
\rowcolor{myBlue!60}   \textit{w/ \method} & 5M & \ding{52}  & 78.94&82.75 &85.91	& 6.46  & 19.73	& 26.45 &51.26	& 45.92  \\

ChameleonVQ & - &\ding{55}  &58.75&62.65  & 81.59&	7.94  &10.12	&13.73 & 49.71&	46.34	\\
\rowcolor{myBlue!60}   \textit{w/ \method} & 5M & \ding{52}  & 72.16&76.39 &83.70	& 7.20  & 14.59	&19.80&	50.20	&46.02	\\

\addlinespace[2pt]
\hline
\rowcolor{grpgray}\multicolumn{11}{c}{{\textit{High Resolution}}} \\

\addlinespace[2pt]

LlamagenVQ & 50M & \ding{55}  & 92.04	& 92.45 & 92.62&	5.67  & 33.47	&40.98 & 52.50	&42.55	\\
\rowcolor{myBlue!60} \textit{w/ \method} & 5M& \ding{52}  & 93.39	&92.42 & 93.32&	5.60  &42.15&	47.72 &	53.29	&42.55		\\

ChameleonVQ & - &\ding{55}  & 91.70	& 91.49 & 92.25&	5.10  & 30.49	&36.31&51.79	&42.11	\\
\rowcolor{myBlue!60} \textit{w/ \method} & 5M & \ding{52}  & 92.19	&91.78 & 92.21	&4.62  &36.47	&42.29 &	52.12	&42.14 	\\
\bottomrule

\end{tabular}

\label{tab:sub_tokenzier_add_tokenizer}
\end{table*}





\subsection{AR Model Results on TextAtlasEval}
We also evaluate general AR models on the TextAtlasEval. The results are reported in ~\cref{tab:sup_ar_model_textatlas}.

Across different AR backbones and resolutions, RDA provides consistent improvements on TextAtlasEval benchmarks.

\section{More Visualizations}

\subsection{Tokenizer Reconstruction Results}

\begin{figure*}[t]
  \centering
  \includegraphics[width=1\textwidth]{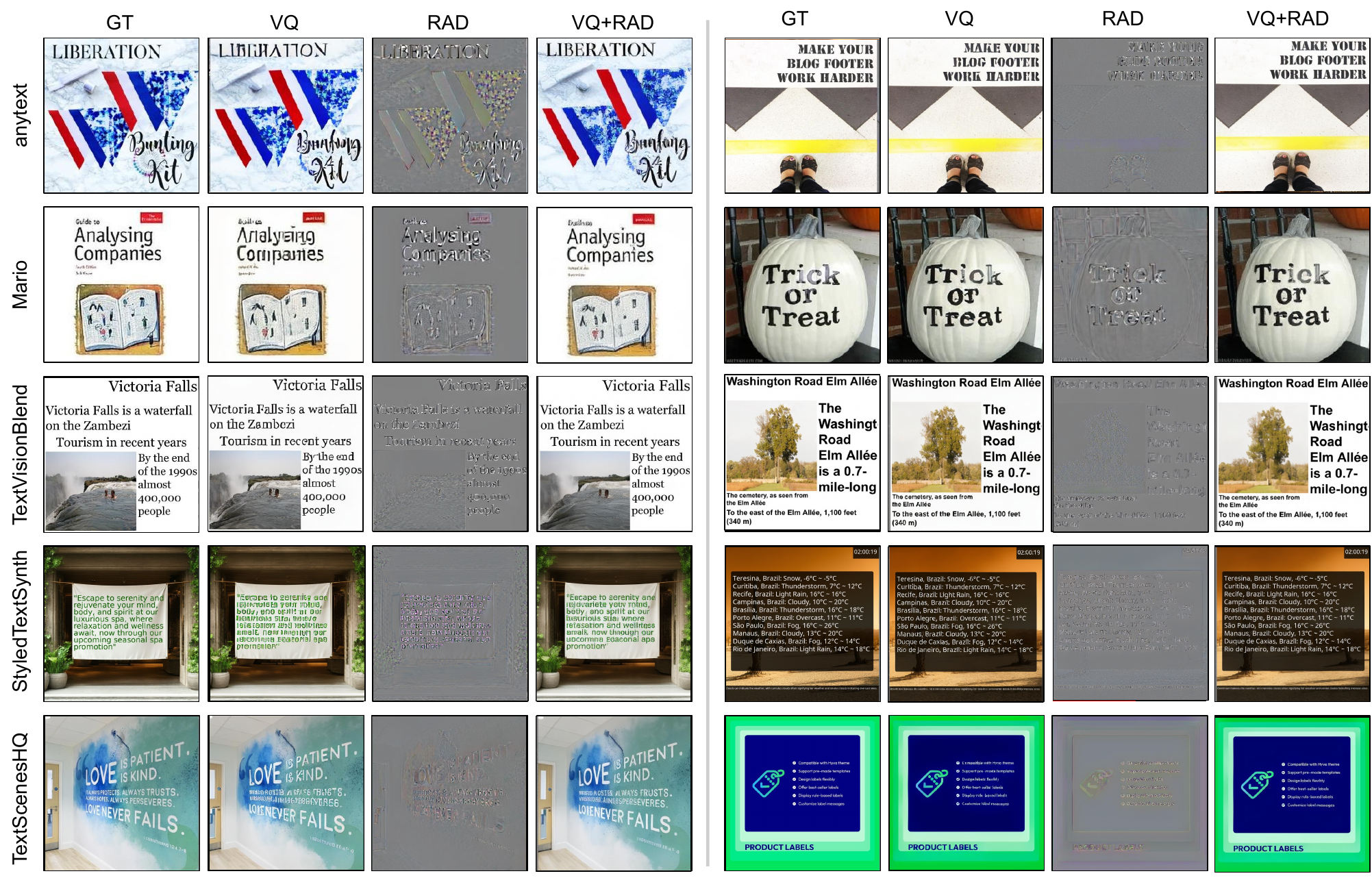} 
  \caption{\textbf{Qualitative Results of LlamaGenVQ Applying RDA.} Left: low-resolution setting. Right: high-resolution setting.}
  \label{fig:sub_vis_llamagenvq}
\end{figure*}

\begin{figure*}[t]
  \centering
  \includegraphics[width=1\textwidth]{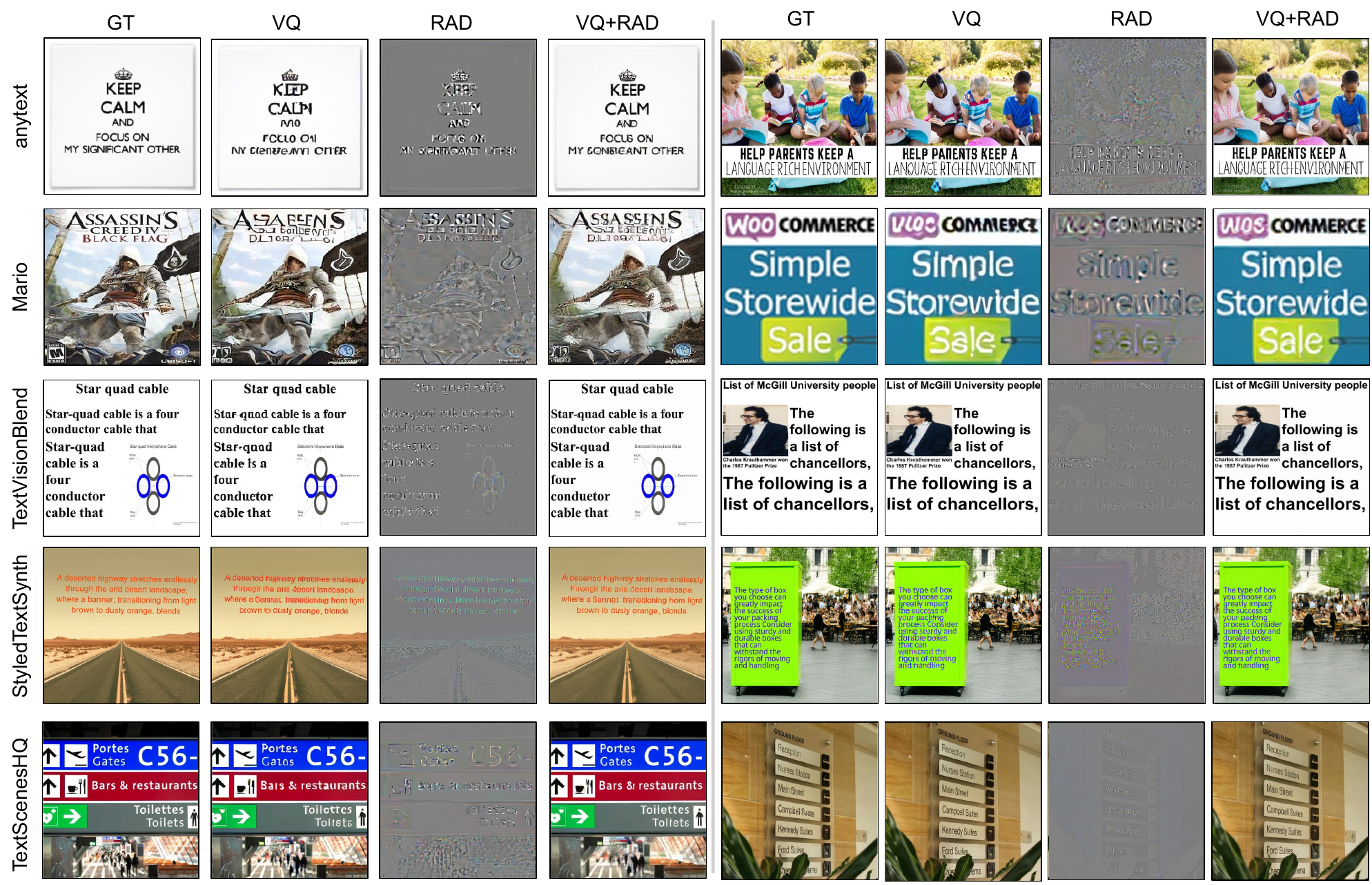} 
  \caption{\textbf{Qualitative Results of ChameleonVQ Applying RDA.} Left: low-resolution setting. Right: high-resolution setting.}
  \label{fig:sub_vis_chameleonvq}
\end{figure*}

We evaluate the reconstruction quality of different image tokenizers, including LlamaGen-VQ and Chameleon-VQ~\cite{chameleon}, across multiple datasets. For clarity, ~\cref{fig:sub_vis_llamagenvq} presents results based on LlamaGenVQ, while ~\cref{fig:sub_vis_chameleonvq} shows results from Chameleon-VQ. Across both tokenizers, applying RDA leads to visibly sharper text strokes and improved structural consistency, demonstrating that our method generalizes to different tokenizer architectures.

\begin{table*}[t]
\centering
\footnotesize
\caption{\textbf{Comparison of general AR models before and after applying \method.} Each cell shows the result  \dualcolor{\textit{\textbf{w}}}{\textit{\textbf{wo}}}  applying RDA. }
\resizebox{\textwidth}{!}{

\begin{tabular}{l|r|r|ccc|ccc|ccc}
\toprule
\multirow{2}{*}{\centering \textbf{Model}} &
\multirow{2}{*}{\centering \textbf{Size}} &
\multirow{2}{*}{\centering \textbf{Res}} &
\multicolumn{3}{c|}{\textbf{TextVisionBlend}} &
\multicolumn{3}{c|}{\textbf{StyledTextSynth}} &
\multicolumn{3}{c}{\textbf{TextScenesHQ}}\\
& & &Acc.~$\uparrow$ & F1.~$\uparrow$ & CER~$\downarrow$ & Acc.~$\uparrow$ & F1.~$\uparrow$ & CER~$\downarrow$
& Acc.~$\uparrow$ & F1.~$\uparrow$ & CER~$\downarrow$ \\
\midrule


\multirow{2}{*}{\makecell[l]{Janus Pro}}
& 1B & 384  &
\dualcolor{0.68}{\textbf{0.74}} &
\dualcolor{1.22}{\textbf{1.30}} &
\dualcolor{0.94}{\textbf{0.93}} &
\dualcolor{0.47}{\textbf{0.64}} &
\dualcolor{0.91}{\textbf{1.21}} &
\dualcolor{0.97}{0.97} &
\dualcolor{0.39}{\textbf{1.02}} &
\dualcolor{0.70}{\textbf{1.70}} &
\dualcolor{0.94}{\textbf{0.90}}
\\

& 7B & 384  &
\dualcolor{0.37}{\textbf{0.56}} &
\dualcolor{0.69}{\textbf{1.04}} &
\dualcolor{0.97}{\textbf{0.96}} &
\dualcolor{0.46}{\textbf{0.89}} &
\dualcolor{0.60}{\textbf{1.14}} &
\dualcolor{0.98}{\textbf{0.97}} &
\dualcolor{0.68}{\textbf{1.02}} &
\dualcolor{1.19}{\textbf{1.70}} &
\dualcolor{0.92}{\textbf{0.90}}
\\

\addlinespace[2pt]
\hline
\addlinespace[2pt]
\multirow{4}{*}{TAR}
& 1.5B & 512  &
\dualcolor{0.87}{\textbf{1.32}} &
\dualcolor{1.55}{\textbf{2.26}} &
\dualcolor{0.96}{\textbf{0.95}} &
\dualcolor{1.17}{\textbf{2.02}} &
\dualcolor{2.15}{\textbf{3.55}} &
\dualcolor{0.96}{\textbf{0.94}} &
\dualcolor{2.31}{\textbf{3.59}} &
\dualcolor{3.59}{\textbf{5.04}} &
\dualcolor{0.89}{\textbf{0.87}} \\
& 1.5B & 1024 &
\dualcolor{2.81}{\textbf{4.05}} &
\dualcolor{4.00}{\textbf{4.57}} &
\dualcolor{0.92}{0.92} &
\dualcolor{3.48}{\textbf{5.10}} &
\dualcolor{5.52}{\textbf{7.10}} &
\dualcolor{0.89}{\textbf{0.86}} &
\dualcolor{2.31}{\textbf{3.59}} &
\dualcolor{3.59}{\textbf{5.04}} &
\dualcolor{0.89}{\textbf{0.85}} \\
& 7B & 512 &
\dualcolor{4.90}{\textbf{7.29}} &
\dualcolor{7.34}{\textbf{10.11}} &
\dualcolor{0.87}{\textbf{0.85}} &
\dualcolor{3.62}{\textbf{6.39}} &
\dualcolor{6.21}{\textbf{9.99}} &
\dualcolor{0.92}{\textbf{0.88}} &
\dualcolor{9.84}{\textbf{13.63}} &
\dualcolor{13.31}{\textbf{16.61}} &
\dualcolor{0.77}{\textbf{0.74}} \\
& 7B & 1024 &
\dualcolor{7.56}{\textbf{7.98}} &
\dualcolor{10.33}{\textbf{10.74}} &
\dualcolor{0.84}{0.84} &
\dualcolor{7.70}{\textbf{8.32}} &
\dualcolor{11.70}{\textbf{12.40}} &
\dualcolor{0.86}{\textbf{0.85}} &
\dualcolor{13.94}{\textbf{14.47}} &
\dualcolor{16.59}{\textbf{16.83}} &
\dualcolor{0.75}{0.75} \\
\bottomrule

\end{tabular}
}

\label{tab:sup_ar_model_textatlas}
\end{table*}




\subsection{General AR Generation Results}
\begin{figure*}[t]
  \centering
  \includegraphics[width=1\textwidth]{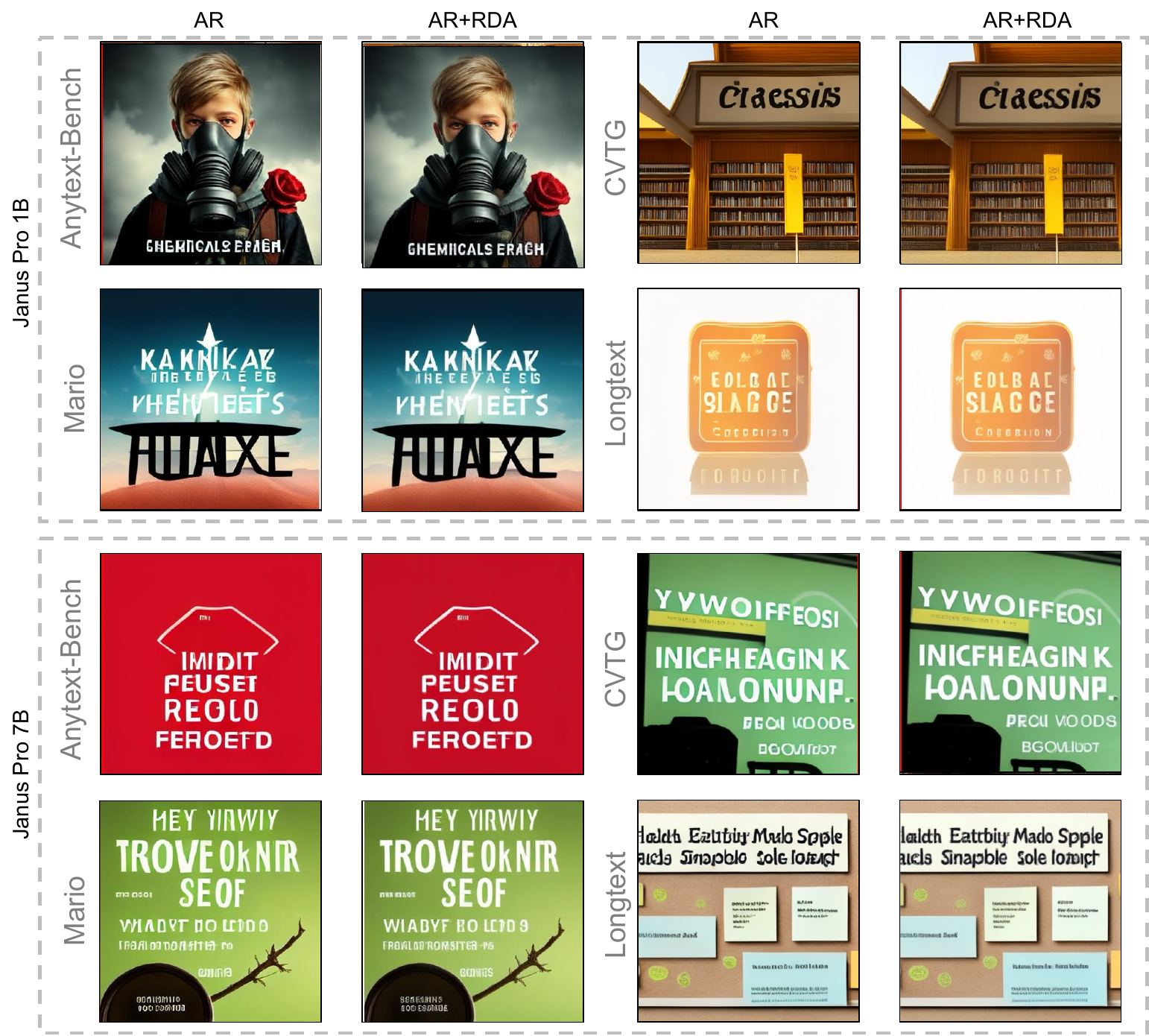} 
  \caption{\textbf{Qualitative Results of Janus Pro Applying RDA}}
  \label{fig:sub_vis_janus}
\end{figure*}

\begin{figure*}[t]
  \centering
  \includegraphics[width=1\textwidth]{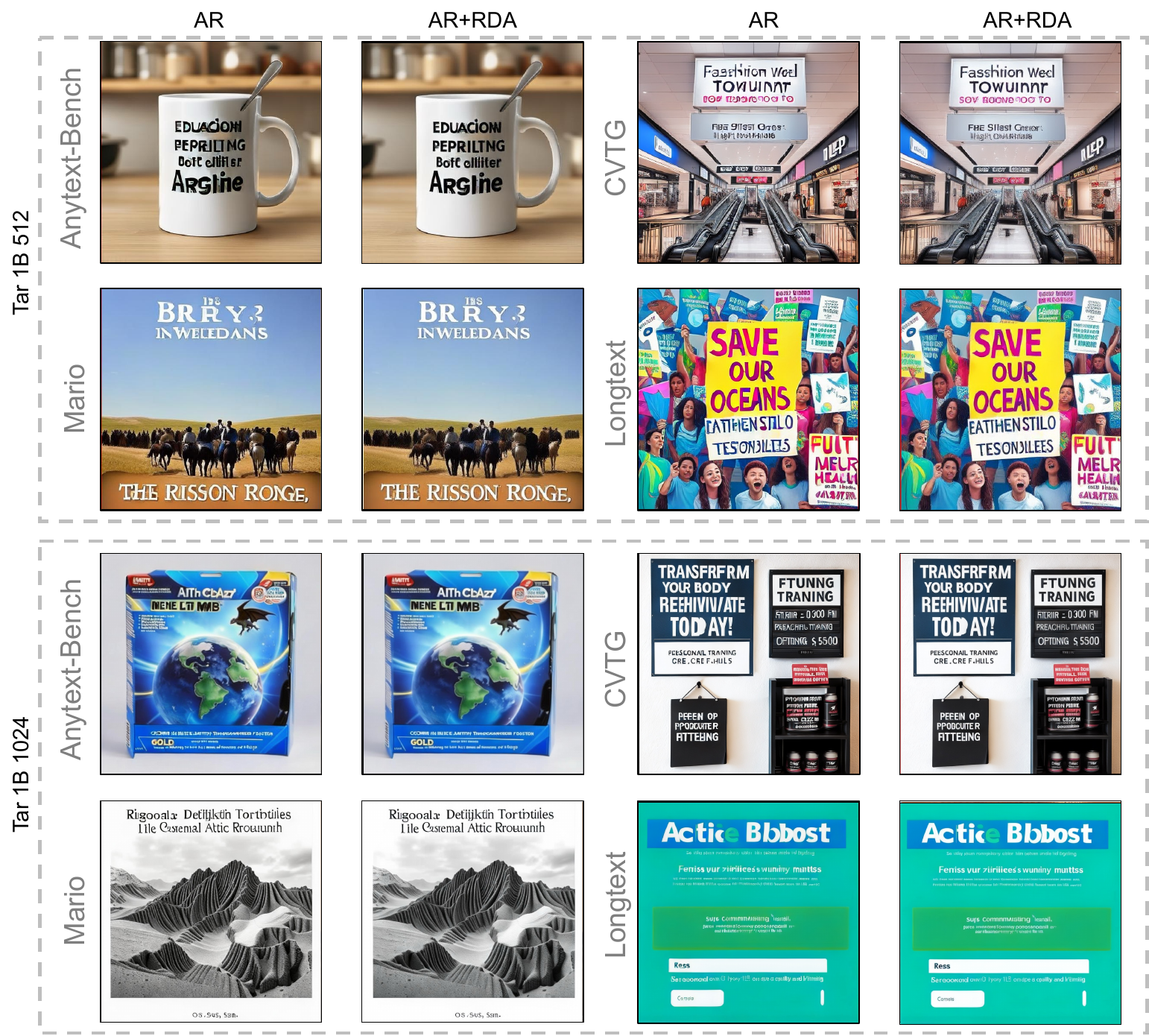} 
  \caption{\textbf{Qualitative Results of Tar 1B Applying RDA}}
  \label{fig:sub_vis_tar1b}
\end{figure*}

\begin{figure*}[t]
  \centering
  \includegraphics[width=1\textwidth]{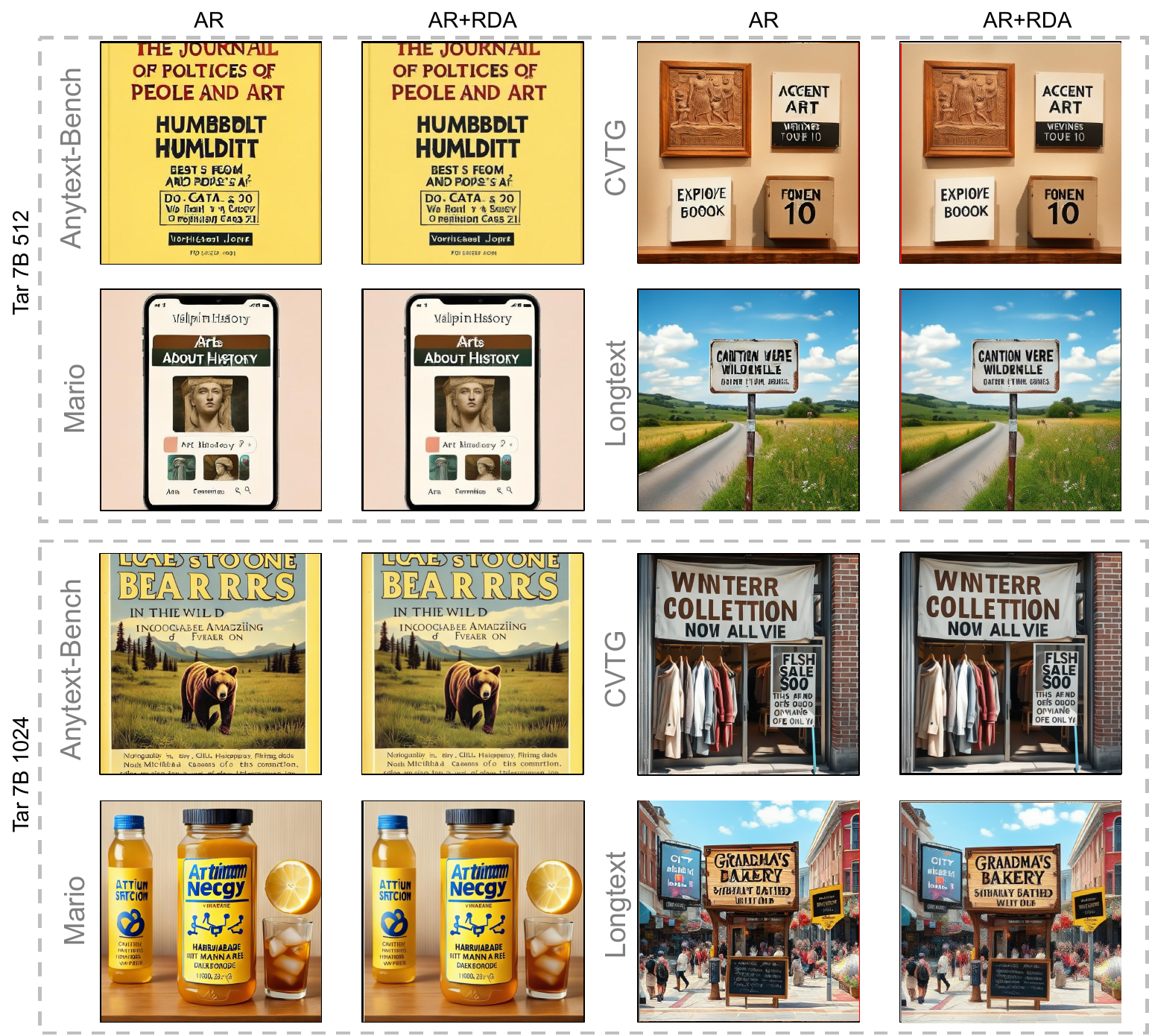} 
  \caption{\textbf{Qualitative Results of Tar 7B Applying RDA}}
  \label{fig:sub_vis_tar7b}
\end{figure*}

\textbf{Janus Pro} ~\cref{fig:sub_vis_janus} shows generation results from Janus Pro.
\textbf{TAR-1B} ~\cref{fig:sub_vis_tar1b} shows generation results from TAR-1B.
\textbf{TAR-7B} ~\cref{fig:sub_vis_tar7b} shows generation results from TAR-7B.

\subsection{Text-Specialized AR Generation Results}
\begin{figure*}[t]
  \centering
  \includegraphics[width=1\textwidth]{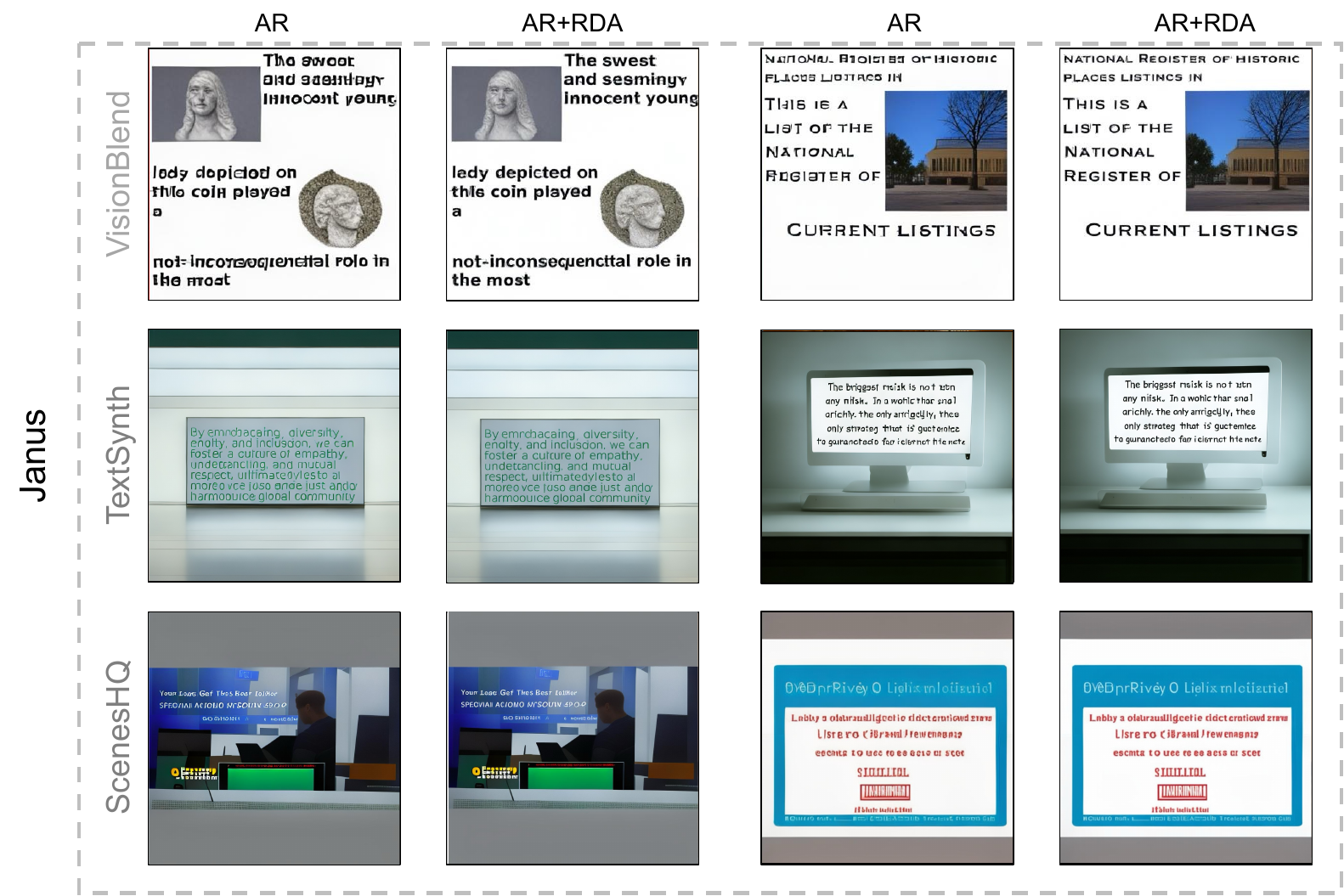} 
  \caption{\textbf{Qualitative Results of Finetuned Janus Pro Applying RDA}}
  \label{fig:sub_vis_trained_janus}
\end{figure*}

\textbf{Janus Pro (Fine-tuned)} ~\cref{fig:sub_vis_trained_janus} shows generation results from fine-tuned Janus Pro.

\begin{figure*}[t]
  \centering
  \includegraphics[width=1\textwidth]{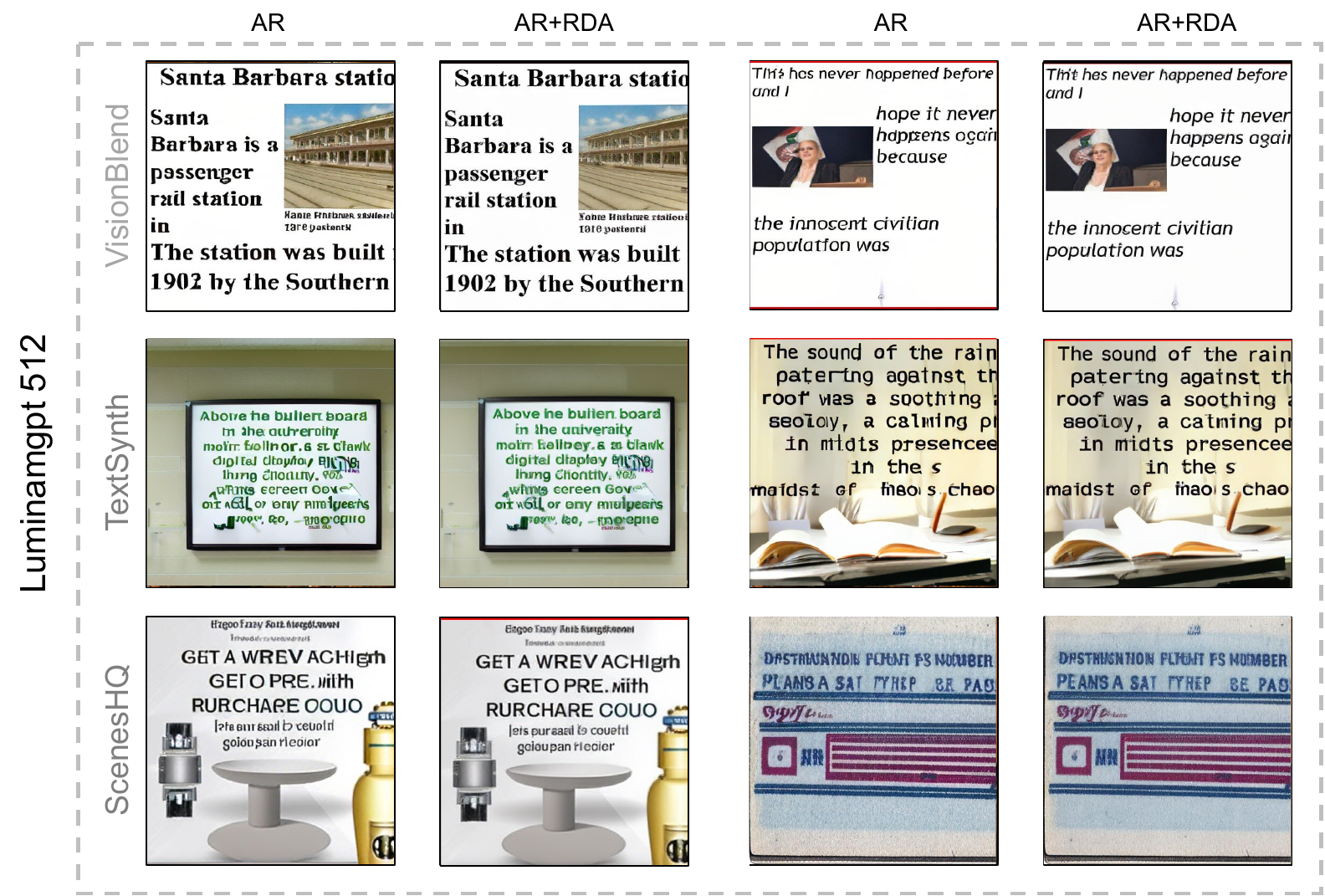} 
  \caption{\textbf{Qualitative Results of Finetuned Luminamgpt 512 Applying RDA}}
  \label{fig:sub_vis_trained_lumina_512}
\end{figure*}

\textbf{Lumina-mGPT (512px)}~\cref{fig:sub_vis_trained_lumina_512} shows generation results fine-tuned Lumina-mGPT at 512 resolution.

\begin{figure*}[t]
  \centering
  \includegraphics[width=1\textwidth]{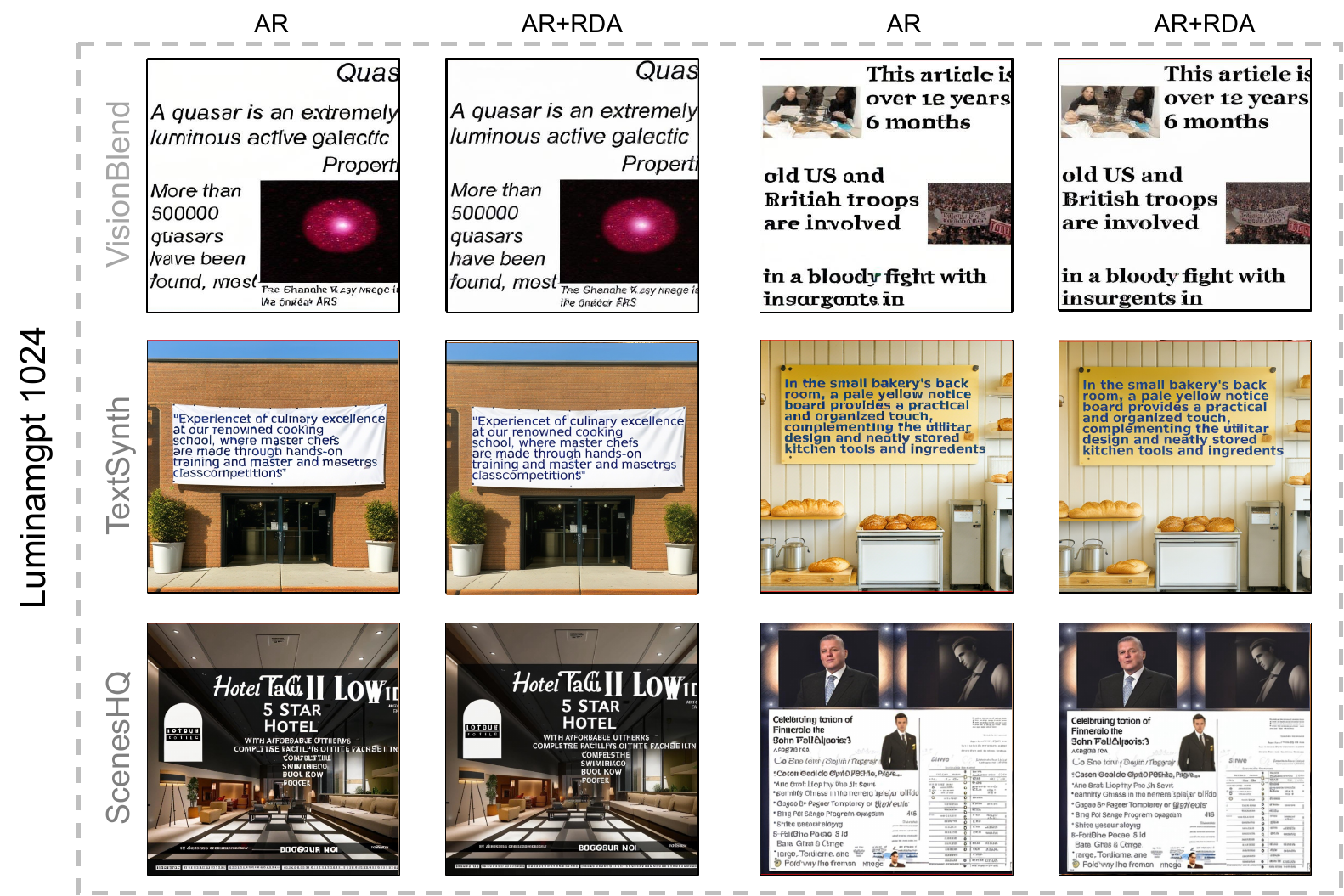} 
  \caption{\textbf{Qualitative Results of Finetuned Luminamgpt 1024 Applying RDA}}
  \label{fig:sub_vis_trained_lumina_1024}
\end{figure*}

\textbf{Lumina-mGPT (1024px)}~\cref{fig:sub_vis_trained_lumina_1024} shows generation results fine-tuned Lumina-mGPT at 1024 resolution.


\section{Additional Information}
\subsection{Recaption Prompt}
We use the following prompt to generate recaptions via Qwen-2.5-VL:
\begin{prompt}{Recaption Prompt}
Carefully describe the image by precisely combining visual elements with all visible text.  The final caption must integrate the visual scene and quoted text into a coherent, factual narrative of around 100 words. Extract every piece of visible text from the image—no omissions—and enclose each text string in double quotes (""). Specify the approximate position of each text (e.g., top-left, center, bottom-right). Avoid adding any imaginative, inferred, or generic descriptions not grounded in the image.
\end{prompt}


\end{document}